\documentclass{article}

\PassOptionsToPackage{numbers,compress}{natbib}

\usepackage[preprint]{main}

\usepackage[utf8]{inputenc}
\usepackage[T1]{fontenc}
\usepackage[hypertexnames=false]{hyperref}
\hypersetup{hidelinks}
\usepackage{url}
\usepackage{xurl}
\usepackage{booktabs}
\usepackage{amsfonts}
\usepackage{nicefrac}
\usepackage{microtype}
\usepackage[table]{xcolor}
\definecolor{ClaudeOrange}{HTML}{C2714F}
\definecolor{ClaudeDark}{HTML}{3D3929}
\definecolor{ClaudeBg}{HTML}{FAF9F7}
\definecolor{WarmGray}{HTML}{8C8577}
\newcommand{\hl}[1]{\colorbox{ClaudeOrange!15}{#1}}
\usepackage{tabularx}
\usepackage{float}
\usepackage{graphicx}
\usepackage{subcaption} %
\usepackage{wrapfig}
\usepackage{listings}
\usepackage{tcolorbox}
\tcbuselibrary{listings, skins, breakable}
\newtcblisting[auto counter]{promptbox}[2][]{%
  enhanced jigsaw,
  listing only,
  listing options={
    escapeinside={(*@}{@*)},
    basicstyle=\ttfamily\small,
    breaklines=true,
    breakatwhitespace=true,
    breakindent=0pt,        %
    breakautoindent=false,  %
    columns=fullflexible,
    keepspaces=true,
    tabsize=2,
    showspaces=false,
    showstringspaces=false,
    showtabs=false,
    upquote=true,
    aboveskip=0pt,          %
    belowskip=0pt,          %
  },
  colback=ClaudeBg,
  colframe=ClaudeDark,
  coltitle=white,
  colbacktitle=ClaudeDark,
  fonttitle=\bfseries,
  title={Listing~\thetcbcounter: #2},
  boxrule=0.5pt,
  borderline west={2.5pt}{0pt}{ClaudeOrange},
  arc=2pt,
  boxsep=0pt,                                    %
  left=8pt, right=8pt, top=-12pt, bottom=4pt,
  toptitle=3pt, bottomtitle=3pt,
  attach boxed title to top={yshift=0pt},
  boxed title style={colframe=ClaudeDark, colback=ClaudeDark, boxrule=0pt, sharp corners=downhill, arc=2pt},
  before skip=8pt, after skip=8pt,
  #1
}
\BeforeBeginEnvironment{lstlisting}{%
  \setlength{\topsep}{0pt}%
  \setlength{\partopsep}{0pt}%
  \setlength{\parskip}{0pt}%
}
\usepackage[all]{hypcap}
\usepackage{algorithm}
\usepackage{algorithmic}
\usepackage{amsmath}
\usepackage{amsthm}

\newcommand\our{\makebox{\textsc{ATLAS}}}
\newcommand\attsmm{\makebox{\textsc{ATLAS-MM}}}
\newcommand\attsmi{\makebox{\textsc{ATLAS-MI}}}
\newcommand\explore{\texttt{explore}}
\newcommand\integrate{\texttt{integrate}}

\newcommand{\groupheader}[2]{\rowcolor{WarmGray!20}\multicolumn{#1}{c}{\textit{#2}}\\}

\makeatletter
\newcommand{\captionof}[1]{\def\@captype{#1}\caption}
\makeatother

\newcommand{\refparagraph}[2]{\phantomsection\label{#1}\paragraph{#2}}

\title{ATLAS: Agentic Test-time Learning-to-Allocate Scaling}
\author{%
  Peijia Qin \\
  University of California, San Diego \\
  \texttt{pqin@ucsd.edu} \\
  \And
  Qi Cao \\
  University of California, San Diego \\
  \texttt{q9cao@ucsd.edu} \\
  \And
  Pengtao Xie \\
  University of California, San Diego \\
  \texttt{p1xie@ucsd.edu} \\
}

\begin{document}

\raggedbottom

\maketitle

\begin{abstract}
Test-time scaling has become a major way to improve large language model reasoning, but its orchestration has remained designer-engineered: a fixed sample budget, a fixed refinement loop, a fixed scoring rule, or a fixed search policy decides how compute is spent, leaving the model in charge of solving but not of orchestration. We introduce \our{}, an agentic test-time scaling framework in which an LLM orchestrator owns the control loop end-to-end. Through a single action, \explore{}, which dispatches a fresh independent solver on the original problem, the orchestrator decides whether to gather more evidence, when to stop, and how to synthesize the final answer; the action space is extensible, with each \explore{} call optionally specifying solver, reasoning effort, or prompting strategy. We evaluate \our{} on four benchmarks covering scientific question answering, code generation, and multimodal reasoning under a Claude Sonnet 4.6 backbone, where it reaches $56.00\%$ on HLE-Verified, $82.29\%$ on LiveCodeBench, $85.86\%$ on GPQA-Diamond, and $23.71\%$ on BabyVision while using far fewer API calls than fixed-workflow baselines. A multi-model extension, \attsmm{}, that exposes solver choice as an additional action dimension further improves HLE-Verified to $60.00\%$ and LiveCodeBench to $85.63\%$, with consistent gains on GPQA-Diamond and BabyVision. Ablations replacing the orchestrator's direct synthesis with a separate integrator degrade or fail to improve accuracy on three of four benchmarks, consistent with the role of \emph{stateful evidence management} in producing the gains.
\end{abstract}

\section{Introduction}

Scaling inference-time compute improves large language model reasoning, but how to allocate that compute remains an open problem. No single strategy (sequential, parallel, or hybrid) is uniformly optimal across problems of varying difficulty \citep{snell_test_time_scaling_paper}, and naive scaling can saturate or even degrade accuracy through poor coverage and reward hacking \citep{best_of_n_optimality_paper}. The central question is not \emph{whether} to scale, but \textbf{who orchestrates the scaling}: a designer in advance, or the model itself in context.

Test-time scaling has progressed mainly by enriching the solving step (more samples, longer chains, deeper trees, learned verifiers), while the orchestration of that compute has remained \textbf{designer-engineered}. Self-Refine encodes a fixed critique-revise loop \citep{self_refine_paper}. Best-of-$N$ reranking encodes a fixed sampling count and a fixed scoring rule \citep{snell_test_time_scaling_paper,skywork_reward_v2_paper}. Budget Forcing encodes a fixed token cap \citep{s1_budget_forcing_paper}. Tree of Thoughts and step-level process-reward frameworks encode a fixed expansion-and-pruning policy \citep{tree_of_thoughts_paper,math_shepherd_paper}. Adaptive-allocation methods soften the rigidity by tying the stop decision to vote distributions or scalar reward scores \citep{adaptive_consistency_paper,damani_adaptive_allocation_paper}. The rule itself, however, is still external to the model. Recent agentic methods bring LLMs into the loop yet typically reserve the model for solving while a hand-specified controller decides when to stop \citep{catts_webagents_paper,tumix_paper}.

\begin{figure}[t]
  \centering
  \vspace{-1em}
  \includegraphics[width=\textwidth]{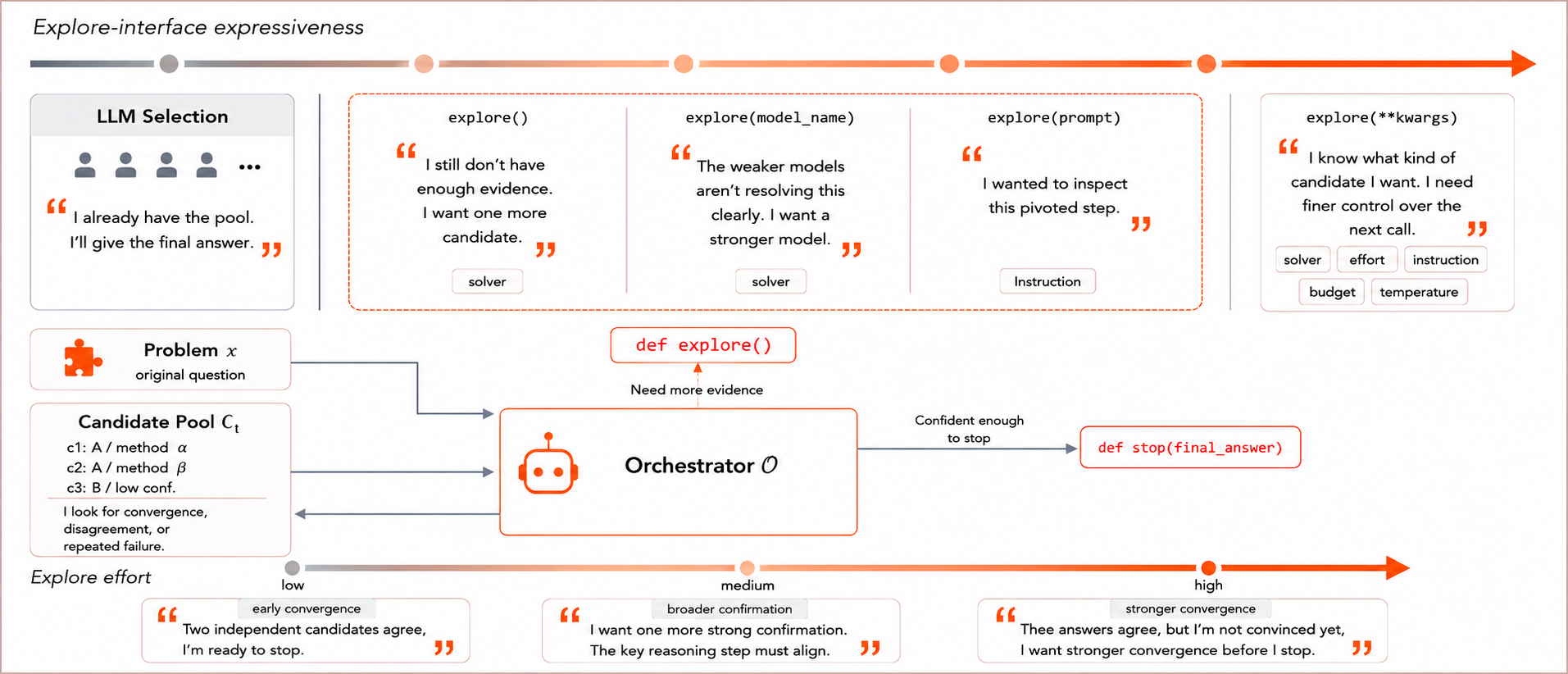}
  \caption{\our{} casts test-time scaling as adaptive action selection over an extensible \explore{} action space. The orchestrator observes the original problem and accumulated candidate pool, decides whether more evidence is needed, dispatches fresh independent solver calls through \explore{}, and stops once the evidence is sufficient to synthesize a final answer. A richer action space exposes additional control dimensions such as solver choice, reasoning effort, prompting strategy, budget, or sampling configuration.}
  \label{fig:atlas-action-space}
\end{figure}

We introduce \our{} (\textbf{A}gentic \textbf{T}est-time \textbf{L}earning-to-\textbf{A}llocate \textbf{S}caling), where \textbf{the model itself}, not a hand-specified rule, schedule, or scoring function, decides how much compute to spend on each problem and when to stop. As illustrated in Figure~\ref{fig:atlas-action-space}, an LLM orchestrator manages a growing pool of candidate solutions through a single tool, \explore{}, which dispatches a fresh, independent solver on the original problem. Each \explore{} call sees only the original problem, so candidates are independent; only the orchestrator accumulates evidence across calls, and its stop and synthesis decisions both rest on this single stateful in-context view. The action space is extensible: each \explore{} call can in principle specify which solver to invoke, at what reasoning effort, and with what decomposition strategy. We evaluate the simplest instantiation, a single parameterless tool whose only decision is when to stop, and find that even this minimal configuration achieves competitive accuracy at substantially lower cost. \attsmm{}, a multi-model extension that adds solver choice to the action space, further reaches $60.00\%$ on HLE-Verified and $85.63\%$ on LiveCodeBench, with consistent gains on GPQA-Diamond and BabyVision, supporting the principle that the orchestrator's value scales with the richness of its decision space.
We evaluate \our{} across four benchmarks covering scientific question answering, code generation, and multimodal reasoning: HLE-Verified \citep{hle_verified_paper}, LiveCodeBench \citep{livecodebench_paper}, BabyVision \citep{babyvision_paper}, and GPQA-Diamond \citep{gpqa_paper}. Because the orchestrator's prompt is benchmark-agnostic, the same controller transfers across text-only and image-conditioned tasks without retraining. Our contributions are as follows:
\begin{enumerate}\setlength{\itemsep}{1pt}\setlength{\parskip}{0pt}
    \item We introduce \our{}, a test-time scaling framework where an LLM orchestrator owns control end-to-end, deciding in context whether to dispatch another fresh solver, when to stop, and how to synthesize. This replaces the hand-specified loops, schedules, or scoring rules that govern prior methods.
    \item We evaluate \our{} on four benchmarks under a Claude Sonnet 4.6 backbone, where it reaches $56.00\%$ on HLE-Verified, $82.29\%$ on LiveCodeBench, $85.86\%$ on GPQA-Diamond, and $23.71\%$ on BabyVision while using far fewer API calls than fixed-workflow baselines. A multi-model extension, \attsmm{}, further reaches $60.00\%$ on HLE-Verified and $85.63\%$ on LiveCodeBench, with consistent gains on GPQA-Diamond and BabyVision.
    \item Through ablations we identify \textbf{stateful evidence management} as a key driver of the gains: replacing the orchestrator's direct synthesis with a separate integrator degrades or fails to improve accuracy on three of four benchmarks.
\end{enumerate}

\section{Related Work}
\label{sec:related-work-conceptually}

\our{} is most related to four areas of prior work. \emph{Test-time scaling} organizes inference-time compute along sequential, parallel, and hybrid axes, covering self-correction, budget-controlled decoding, self-consistency aggregation, and tree search \citep{snell_test_time_scaling_paper,self_refine_paper,s1_budget_forcing_paper,self_consistency_paper,tree_of_thoughts_paper}. \emph{Adaptive compute allocation} studies when to stop sampling and how much to spend per query, via statistical stopping rules, learned difficulty predictors, and rational metareasoning at the token level \citep{adaptive_consistency_paper,damani_adaptive_allocation_paper,rational_metareasoning_llm_paper}. \emph{Agentic orchestration} covers both heuristic schedulers and learned meta-controllers that decide between exploration strategies, including parallel agent mixtures, contextual bandits over high-level moves, and width-versus-depth search \citep{tumix_paper,meta_reasoner_paper,ab_mcts_paper,router_r1_paper}. \emph{Cost-aware agent training} incorporates execution cost into agent objectives, primarily for tool planning, exploration, and step-level credit assignment \citep{catp_llm_paper,calibrate_then_act_paper,toolrl_paper}. Across these four areas, the controller is either hand-specified or separately trained. \our{} stands at a different point: an LLM orchestrator reads the full candidate pool and decides every explore and stop action \textbf{in context}, with no external rule, scoring function, or learned controller in the loop. See Appendix~\ref{app:related-work-extended} for more detailed discussion of related work.

\section{Methodology}
\label{sec:methodology}

\subsection{The \our{} Loop}

The orchestrator repeatedly decides whether to call \explore{} or to stop and synthesize. It \emph{never} solves the problem itself.

\paragraph{Trajectory.} Each turn $t$ produces a structured trace
\begin{tcolorbox}[colback=ClaudeOrange!8, colframe=orange, arc=2pt, boxrule=0.5pt, left=15pt, top=5pt, bottom=5pt]
\small
\begin{tabular}{@{}l@{\quad}l@{}}
\texttt{Thought}$_t$:     & assessment of the candidate pool $\mathcal{C}_{t-1}$ \\
\texttt{Action}$_t$:      & \texttt{explore[\textit{args}]} \quad $|$ \quad \texttt{stop[\textit{final\_answer}]} \\
\texttt{Observation}$_t$: & $c_t = (\text{answer}, \text{reasoning}, \text{approach}, \text{confidence})$ \\
\end{tabular}
\end{tcolorbox}
in which the orchestrator first articulates its assessment, then emits exactly one Action (either \texttt{explore} with arguments determined by the variant in \S\ref{sec:variants}, or \texttt{stop}), and the corresponding Observation is appended to the trajectory. The structure makes every decision auditable and matches the Thought/Action/Observation idiom of agentic systems~\citep{react_paper}.

\begin{wrapfigure}{r}{0.52\textwidth}
\vspace{-1em}
\begin{minipage}{0.52\textwidth}
\begin{algorithm}[H]
\caption{\our{}: Adaptive Test-Time Scaling}
\label{alg:atlas}
\footnotesize
\begin{algorithmic}[1]
\REQUIRE Problem $x$, budget $T$, orchestrator $\mathcal{O}$, solver $\mathcal{S}$
\ENSURE Final answer $y$
\STATE $\tau \leftarrow \emptyset$
\FOR{$t = 1$ \TO $T$}
    \STATE $(\textit{Thought}_t, \textit{Action}_t) \leftarrow \mathcal{O}(x, \tau)$
    \IF{$\textit{Action}_t = \texttt{stop}[y]$}
        \STATE \RETURN $y$
    \ENDIF
    \STATE $c_t \leftarrow \mathcal{S}(x;\, \textit{Action}_t)$
    \STATE $\tau \leftarrow \tau \cup \{(\textit{Thought}_t, \textit{Action}_t, c_t)\}$
\ENDFOR
\STATE $(\_,\, \texttt{stop}[y]) \leftarrow \mathcal{O}(x, \tau)$ \COMMENT{budget exhausted}
\RETURN $y$
\end{algorithmic}
\end{algorithm}
\end{minipage}
\vspace{-1em}
\end{wrapfigure}

\paragraph{The \texttt{explore} Action.} An \texttt{explore} Action dispatches a fresh solver on the original problem $x$. The solver works \emph{from scratch} (it does not see or continue previous candidates), and the returned candidate $c_t = (\text{answer}, \text{reasoning}, \text{approach}, \text{confidence})$ becomes \texttt{Observation}$_t$, appended to $\mathcal{C}_t$. By construction, each explorer receives only the problem statement, guaranteeing that explore results are \emph{conditionally independent} given the problem. This independence lets the orchestrator assess the marginal contribution of each additional explore in isolation.

\paragraph{The \texttt{stop} Action.} A \texttt{stop} Action terminates the loop with a final answer synthesized directly from $\mathcal{C}_{t-1}$ by the orchestrator itself. Because it has observed the entire trajectory (every Thought/Action/Observation triple), it has richer context than a separate aggregator that sees only the final pool. We refer to this property as \emph{stateful evidence management}: the orchestrator's stop and synthesis decisions are both conditioned on the full Thought/Action/Observation history rather than on a final pool of candidates. The orchestrator can issue \texttt{stop} at any point before the explore budget $T$ is exhausted; the decision is made by the orchestrator model itself based on its assessment of the candidate pool. The prompt encodes principles for this judgment: genuine convergence means independent solvers arriving at the same answer through different methods, repeated failures indicate the problem exceeds solver capability, and each additional \texttt{explore} call has a cost that must be justified by expected information gain. This judgment reflects the cost-benefit tradeoff formalized by \emph{rational metareasoning} and the \emph{value-of-computation} criterion~\citep{russell1991right,hay_selecting_computations}: additional exploration is warranted only when the expected information gain justifies the cost.

Algorithm~\ref{alg:atlas} summarizes the procedure. The control logic is benchmark-agnostic: only the explorer prompt and answer schema change across tasks. The orchestrator prompt encodes its decision logic as declarative principles (e.g.\ ``a single candidate does not constitute sufficient evidence'') rather than prescriptive rules, paired with worked examples that demonstrate how these principles apply in concrete scenarios. Appendix~\ref{app:implementation-details} gives the full implementation details including representative prompts.

\subsection{Action Space and Variants}
\label{sec:extensible-action-space}
\label{sec:variants}

The value of the framework scales with the richness of the \explore{} action space. In the minimal instantiation \texttt{explore[]} takes no arguments and the orchestrator can only decide \emph{when} to stop. Enriching the action space opens additional decision axes, turning the orchestrator into a controller that adaptively selects \emph{what} to compute. Table~\ref{tab:explore-action-space} summarizes the three variants we consider. All variants share the same Thought/Action/Observation trajectory; only the argument field of the \explore{} action changes.

\begin{table}[H]
\centering
\caption{\explore{} action-space variants. All variants use the same Thought/Action/Observation loop; only the argument field exposed by \explore{} changes.}
\label{tab:explore-action-space}
\setlength{\tabcolsep}{4pt}
\renewcommand{\arraystretch}{1.2}
\begin{tabular}{@{}l l l l@{}}
\toprule
\textbf{Variant} & \textbf{\explore{} action} & \textbf{Argument domain} & \textbf{Decision exposed} \\
\midrule
\our{} & \texttt{explore[]} & $\emptyset$ & stop timing \\
\attsmm{} & \texttt{explore[model\_name=$M_i$]} & $M_i \in \{\mathcal{M}_1, \dots, \mathcal{M}_K\}$ & solver capability \\
\attsmi{} & \texttt{explore[instruction=$s$]} & $s \in \text{String}$ & solver focus \\
\bottomrule
\end{tabular}
\end{table}

The two augmented variants address different failure modes: \attsmm{} targets \emph{capability} ceilings by escalating across solver tiers, and \attsmi{} targets \emph{focus} failures by directing a fresh solver toward a localizable disagreement. Both share the same loop, candidate pool, stopping logic, and direct orchestrator synthesis, so the same core policy addresses different failure modes without architectural complexity.

\subsection{Explore Effort}
\label{sec:explore-effort}

Beyond the choice of variant, the orchestrator faces an independent decision: at what point does the accumulated evidence justify commitment? Through the lens of \citet{hay_selecting_computations}'s value-of-computation criterion (\S\ref{sec:methodology}), the orchestrator should stop precisely when the expected information gain from another \texttt{explore} no longer justifies its cost; what counts as ``sufficient gain'' is itself a tunable disposition rather than a universal constant. We expose this disposition as an explicit hyperparameter, \emph{explore effort}, with three discrete levels that set the orchestrator's stopping bar independently of which variant in \S\ref{sec:variants} is being run.

Table~\ref{tab:explore-effort-levels} contrasts the three levels along several lenses.

\begin{table}[H]
\centering
\caption{Explore-effort levels viewed through several lenses.}
\label{tab:explore-effort-levels}
\small
\setlength{\tabcolsep}{4pt}
\renewcommand{\arraystretch}{1.2}
\begin{tabular}{@{}l l l l@{}}
\toprule
 & \multicolumn{1}{c}{\shortstack{\textbf{Low}\\[3pt]\textcolor{orange!35}{\rule{0.8em}{0.35em}}\textcolor{black!8}{\rule{1.6em}{0.35em}}}}
 & \multicolumn{1}{c}{\shortstack{\textbf{Medium}\\[3pt]\textcolor{orange!45}{\rule{1.6em}{0.35em}}\textcolor{black!8}{\rule{0.8em}{0.35em}}}}
 & \multicolumn{1}{c}{\shortstack{\textbf{High}\\[3pt]\textcolor{orange!55}{\rule{2.4em}{0.35em}}}} \\
\midrule
Evidence threshold     & pairwise agreement       & + cross-method confirmation & triangulation or cross-check \\
VoC bar on info gain   & low: weak gain accepted  & moderate                    & high: strong gain required \\
Operating profile      & easy items, fast commit  & mixed difficulty, balanced  & hard items, risk-averse \\
\bottomrule
\end{tabular}
\end{table}

Because explore effort changes only the stopping disposition (not the Action shape the orchestrator can emit), it is fully \emph{orthogonal} to variant choice: any of \our{}, \attsmm{}, \attsmi{} can be operated at any of the three explore-effort levels. The cost-accuracy frontier across the three levels is reported in \S\ref{sec:effort-ablation}.

\section{Experiments}
\label{sec:experiments}

\begin{wrapfigure}{r}{0.48\textwidth}
  \vspace{-5em}
  \centering
  \includegraphics[width=0.48\textwidth]{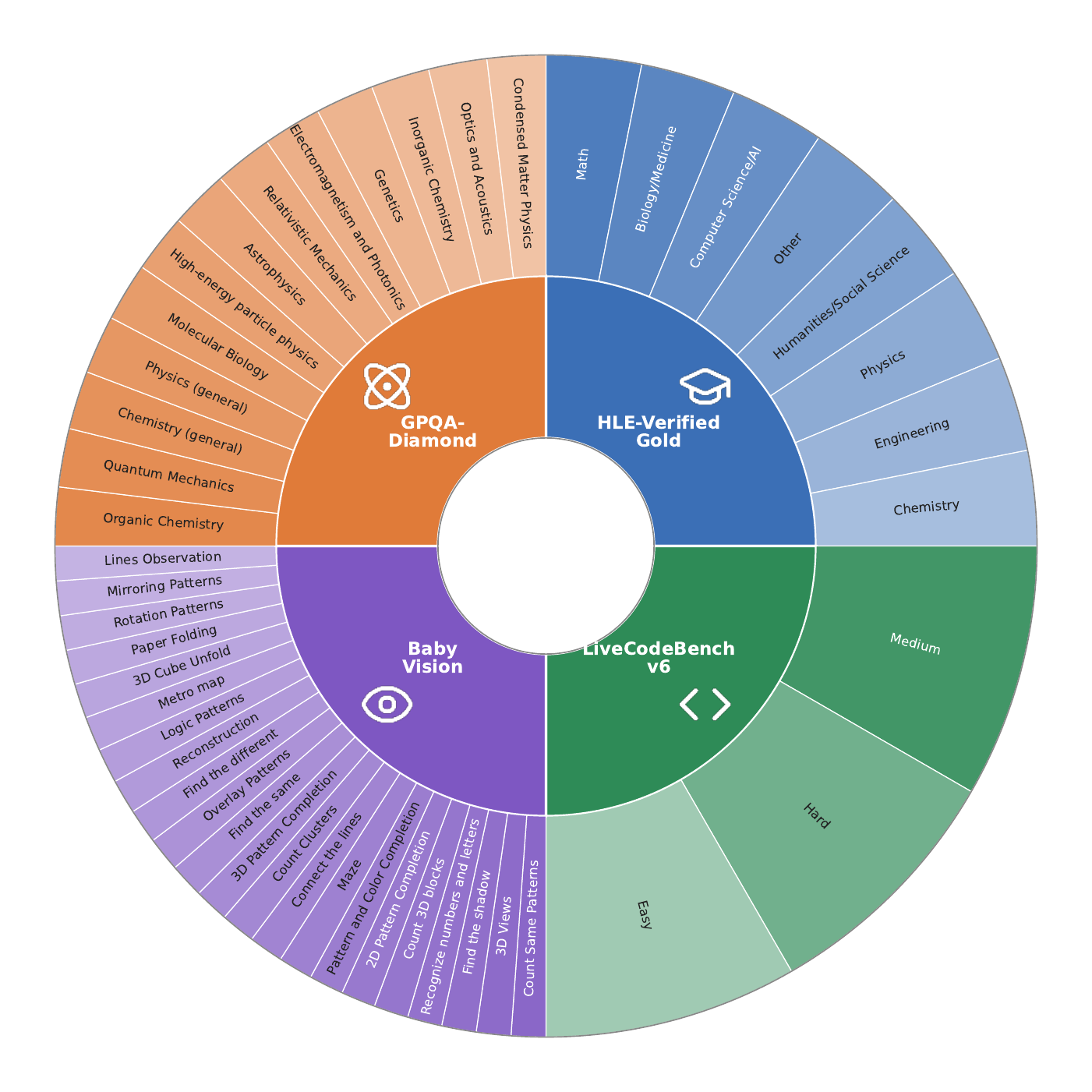}
  \caption{Evaluation coverage. The inner ring shows the four benchmarks; the outer ring expands each into its official subcategories (8 HLE-Verified Gold categories, 3 LiveCodeBench v6 difficulty levels, 22 BabyVision subtypes, 13 GPQA-Diamond subdomains).}
  \label{fig:benchmark-coverage}
  \vspace{-2em}
\end{wrapfigure}
We evaluate \our{} against seven test-time scaling baselines (six in Table~\ref{tab:main-results}; the seventh, LLM Selection, is reported separately in the action-space comparison of Table~\ref{tab:explore-augmentation}) on four benchmarks spanning scientific QA, code generation, and multimodal reasoning (Figure~\ref{fig:benchmark-coverage} summarizes the official subcategories within each). Appendix~\ref{app:implementation-details} and \ref{app:benchmark-details} give the full experimental setups, baseline implementations, and benchmark grading details.

\begin{table}[t!]
\centering
\caption{Main results across four benchmarks. Top group: top-5 zero-shot leaderboard reference per column. Middle group: test-time scaling baselines under Claude Sonnet 4.6. Bottom: \our{} and its multi-model extension \attsmm{} under Claude Sonnet 4.6. \textbf{Bold} marks the best per column among test-time scaling methods. Per-split breakdowns are in Appendix~\ref{app:full-results}.}
\label{tab:main-results}
\small
\setlength{\tabcolsep}{2.5pt}
\renewcommand{\arraystretch}{1.05}
\begin{tabular}{l c c c c}
\toprule
 & HLE-Verified & LiveCodeBench & GPQA-Diamond & BabyVision \\
\midrule
\multicolumn{5}{l}{\textit{Top-5 zero-shot leaderboard reference}} \\
\midrule
1st & \shortstack{52.48\\{\tiny GPT-5.2-H}}      & \shortstack{80.2\\{\tiny o4-mini-H}}     & \shortstack{94.1\\{\tiny Gemini-3.1-Pro}}    & \shortstack{49.7\\{\tiny Gemini-3-Pro}}   \\
2nd & \shortstack{50.16\\{\tiny Opus-4.6}}       & \shortstack{75.8\\{\tiny o3-H}}          & \shortstack{92.0\\{\tiny GPT-5.4-xhi}}       & \shortstack{34.4\\{\tiny GPT-5.2}}        \\
3rd & \shortstack{48.93\\{\tiny Gemini-3-Pro}}   & \shortstack{74.2\\{\tiny o4-mini-M}}     & \shortstack{91.5\\{\tiny GPT-5.3-Codex}}     & \shortstack{30.2\\{\tiny Doubao-1.8}}     \\
4th & \shortstack{48.48\\{\tiny Qwen3-Max-T}}    & \shortstack{73.6\\{\tiny Gemini-2.5-Pro}}& \shortstack{90.8\\{\tiny Gemini-3-Pro}}      & \shortstack{19.2\\{\tiny Qwen3-VL-Plus}}  \\
5th & \shortstack{48.16\\{\tiny Opus-4.5}}       & \shortstack{73.1\\{\tiny DeepSeek-R1}}   & \shortstack{90.3\\{\tiny GPT-5.2-xhi}}       & \shortstack{16.2\\{\tiny Grok-4}}         \\
\midrule
\multicolumn{5}{l}{\textit{Test-time scaling baselines (Claude Sonnet 4.6 backbone)}} \\
\midrule
Pass@1                                              & 48.00          & 77.14          & 83.33          & 19.59          \\
Majority Voting~\citep{self_consistency_paper}      & --             & --             & 83.33          & 23.20          \\
Self-Refine~\citep{self_refine_paper}               & 53.00          & 80.57          & 82.83          & 21.39          \\
Self-Refine (no early stop)~\citep{self_refine_paper} & 58.00          & 82.29          & 85.35          & 21.13          \\
Budget Forcing~\citep{s1_budget_forcing_paper}      & 51.00          & 80.00          & 83.84          & 21.91          \\
Reward-model reranking\textsuperscript{\textdagger} & 52.00          & 78.86          & 83.84          & 20.88          \\
\midrule
\multicolumn{5}{l}{\textit{Our method (Claude Sonnet 4.6 backbone)}} \\
\midrule
\rowcolor{ClaudeOrange!8} \our{}                                                 & 56.00          & 82.29          & 85.86          & 23.71          \\
\rowcolor{ClaudeOrange!8} \attsmm{}                                              & \textbf{60.00} & \textbf{85.63} & \textbf{88.38} & \textbf{23.97} \\
\bottomrule
\end{tabular}

\vspace{2pt}
{\tiny \textsuperscript{\textdagger} Reward-model reranking uses Skywork-Reward-V2 \citep{skywork_reward_v2_paper,skywork_reward_v2_hf} for HLE / LiveCodeBench / GPQA-Diamond and VisualPRM \citep{visualprm_paper,visualprm_hf} for BabyVision.}
\end{table}

\refparagraph{sec:main-results}{Main results.}
Table~\ref{tab:main-results} reports accuracy on all four benchmarks. Table~\ref{tab:explore-augmentation} reports LLM Selection and \attsmm{} as part of the action-space comparison later in this section; per-benchmark scoring conventions and the top/middle/bottom group structure are in Appendix~\ref{app:benchmark-details}; per-split breakdowns are in Appendix~\ref{app:full-results}.

\our{} reaches strong accuracy across all four benchmarks under a Sonnet 4.6 backbone. The multi-model extension \attsmm{} extends this further to $60.00\%$ on HLE-Verified, $85.63\%$ on LiveCodeBench, $88.38\%$ on GPQA-Diamond, and $23.97\%$ on BabyVision -- the column-wise top method on every benchmark in Table~\ref{tab:main-results} and the only row that holds the lead simultaneously on all four, with $+0.77$ to $+3.34$ points over the strongest fixed-budget baseline.

\begin{wraptable}{r}{0.60\textwidth}
  \vspace{-0.8em}
  \centering
  \caption{Controller-side ablations. (a) Orchestrator model effect with Sonnet 4.6 explores throughout. (b) Adding a separate integrator. (c) Disabling extended thinking on GPQA-Diamond.}
  \label{tab:controller-ablations}
  \scriptsize
  \textit{(a) Orchestrator model effect}\\[-1pt]
  \resizebox{\linewidth}{!}{%
  \begin{tabular}{l r r r r r r}
  \toprule
   & \multicolumn{2}{c}{Haiku} & \multicolumn{2}{c}{Sonnet} & \multicolumn{2}{c}{Opus} \\
  \cmidrule(lr){2-3}\cmidrule(lr){4-5}\cmidrule(lr){6-7}
   & Acc. & \$/q & Acc. & \$/q & Acc. & \$/q \\
  \midrule
  HLE-Verified  & 57.00 & 1.83 & 56.00 & 1.59 & 59.00 & 1.50 \\
  LiveCodeBench & 79.43 & 0.71 & 82.29 & 0.53 & 83.43 & 0.64 \\
  GPQA-Diamond  & 82.83 & 0.36 & 85.86 & 0.35 & 86.87 & 0.29 \\
  \bottomrule
  \end{tabular}%
  }
  
  \vspace{0.35em}
  \textit{(b) Separate integrator}\\[-1pt]
  \resizebox{\linewidth}{!}{%
  \begin{tabular}{l r r r r}
  \toprule
   & \multicolumn{2}{c}{\our{}} & \multicolumn{2}{c}{\our{} (+ Integrator)} \\
  \cmidrule(lr){2-3}\cmidrule(lr){4-5}
   & Acc. (\%) & \$/q & Acc. (\%) & \$/q \\
  \midrule
  HLE-Verified  & 56.00 & 1.59 & 57.00 & 2.35 \\
  LiveCodeBench & 82.29 & 0.53 & 80.57 & 0.72 \\
  BabyVision    & 23.71 & 0.27 & 23.71 & 0.38 \\
  GPQA-Diamond  & 85.86 & 0.35 & 85.86 & 0.39 \\
  \bottomrule
  \end{tabular}%
  }

  \vspace{0.35em}
  \scriptsize
  \textit{(c) Extended thinking}\\[-1pt]
  \makebox[\linewidth][c]{%
  \begin{tabular}{l r r}
\toprule
                     & Acc.\ (\%) & \$/q \\
\midrule
\our{}               & 85.86 & 0.35 \\
\our{} (no thinking) & 84.85 & 0.36 \\
\bottomrule
  \end{tabular}%
  }
  \vspace{-1.5em}
  \end{wraptable}
\paragraph{Ablation on the orchestrator.}
We probe three axes of orchestrator capability while holding the explorer (Sonnet 4.6) and prompts fixed: model choice, internal reasoning, and separate integrator. All three ablations isolate \emph{synthesis quality}, not stopping behavior, as the main axis of orchestrator value.

\textit{Model choice} (Table~\ref{tab:controller-ablations}a). Orchestrator accuracy on GPQA-Diamond consistently increases with capability tier: Haiku $82.83\%$ $<$ Sonnet $85.86\%$ $<$ Opus $86.87\%$. Opus also leads on HLE ($59\%$) and LiveCodeBench ($83.4\%$). Haiku is 4 points behind Opus on GPQA-Diamond despite using slightly more explores per question, which isolates aggregation quality rather than effort budget as the main axis.

\textit{Separate integrator} (Table~\ref{tab:controller-ablations}b). By default, \our{} has the orchestrator produce the final answer directly from its accumulated evidence after deciding to stop; \our{} (+ Integrator) replaces this with a dedicated \integrate{} step (a separate model call that receives all candidates and synthesizes the final answer independently). Adding the separate integrator yields accuracy comparable to direct synthesis at higher per-question cost on every benchmark (Table~\ref{tab:controller-ablations}b). \our{} therefore uses direct orchestrator synthesis by default.

\textit{Internal reasoning} (Table~\ref{tab:controller-ablations}c). Disabling extended thinking on a fixed Sonnet 4.6 orchestrator reduces GPQA-Diamond accuracy by about one point ($85.86\% \to 84.85\%$) with identical exploration ($2.14$ vs.\ $2.14$ explores/q). The orchestrator's internal reasoning therefore has a small but positive effect on evidence aggregation.

\begin{wrapfigure}{r}{0.60\textwidth}
  \vspace{-1.0em}
  \centering
  \includegraphics[width=\linewidth]{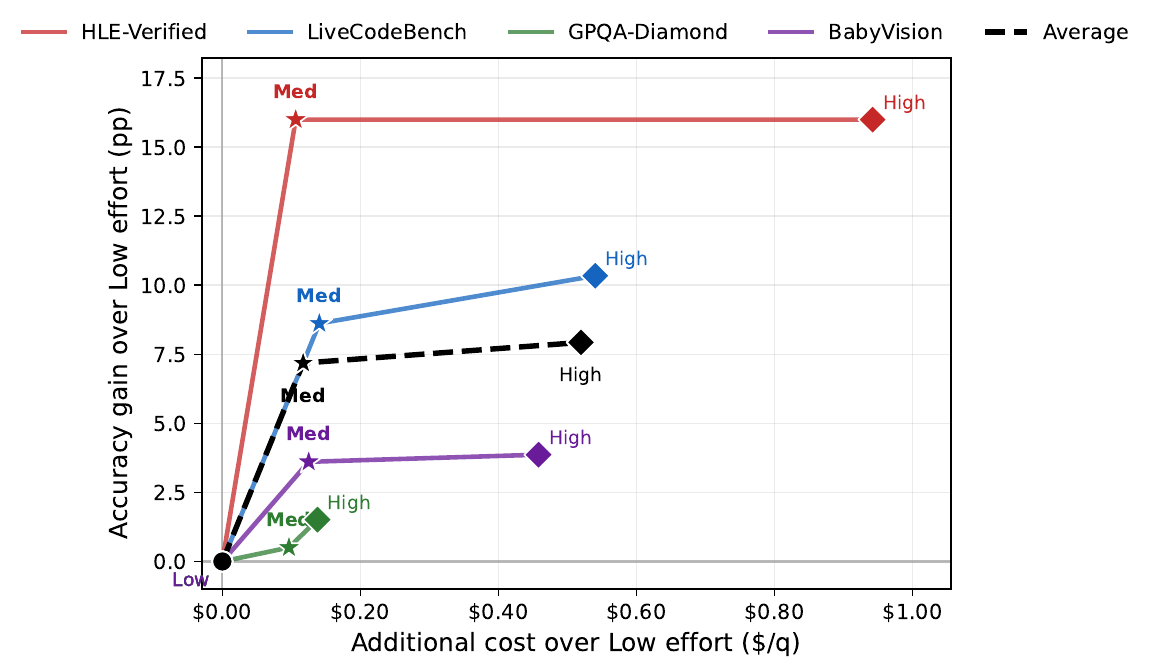}
  \caption{Effect of explore effort on \attsmm{}, plotted as incremental accuracy and cost relative to Low effort. Each curve is one benchmark and each point is an effort level. Higher effort gives non-decreasing accuracy across all benchmarks.}
  \label{fig:effort-ablation}
  \vspace{-1.0em}
\end{wrapfigure}
\paragraph{Ablation on explore effort.}
\label{sec:effort-ablation}
\textit{Aggregate effect} (Figure~\ref{fig:effort-ablation}). Higher effort gives non-decreasing accuracy on every benchmark; cost rises steadily at the Medium-to-High transition, by $1.5${--}$1.8\times$ on the free-form benchmarks (HLE-Verified, LiveCodeBench, BabyVision) and by a milder $+10.5\%$ on GPQA-Diamond.

\begin{figure}[t]
  \centering
  \includegraphics[width=\textwidth]{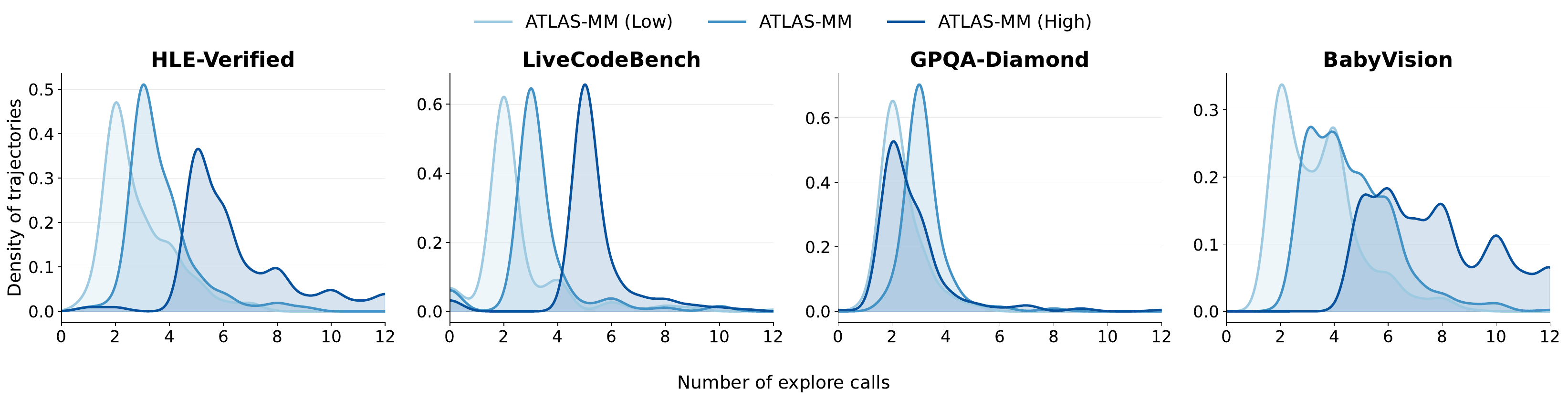}
  \caption{Per-question explore-call distributions for \attsmm{} at the three explore-effort levels (Low / Medium / High), one panel per benchmark. Curves are Gaussian KDEs over the discrete number of explore calls per trajectory. Per-cell sample sizes ($n$) and means ($\mu$), as well as the underlying discrete histograms, appear in Appendix~\ref{app:explore-distributions-effort}.}
  \label{fig:kde-effort-ablation}
\end{figure}

\textit{Per-question distribution} (Figure~\ref{fig:kde-effort-ablation}). On HLE-Verified, LiveCodeBench, and BabyVision the Low$\to$Medium$\to$High transition shifts the distribution rightward at each step, producing the $1.5${--}$1.8\times$ Medium$\to$High cost increase on these three benchmarks. GPQA-Diamond is the exception: at High effort the distribution shifts \emph{leftward} of Medium, because a couple of agreeing explores typically suffice on this highest-accuracy benchmark. The $+10.5\%$ Medium$\to$High cost change (versus $1.5${--}$1.8\times$ on the free-form benchmarks) reflects this fast convergence.

\refparagraph{sec:results-cross-family}{Ablation on the orchestrator backbone.}

The within-family swap above varies the orchestrator across three Anthropic capability tiers while keeping the Sonnet 4.6 explorer fixed. We extend this in two ways, both reported in Table~\ref{tab:cross-family}: (i) an \emph{orchestrator-only swap}, where two open-weights orchestrators from a different model family~\citep{qwen3_paper} are run on top of a Sonnet 4.6 explorer, and (ii) a \emph{full-stack swap}, where the explorer, orchestrator, and synthesizer are all served by the same non-Anthropic backbone. In both setups, prompts and exploration budget ($T{=}8$) are identical to the within-family configuration; only the backbone(s) change. Open-weights backbones are served locally via vLLM; full checkpoint paths and decoding settings are in Appendix~\ref{app:implementation-details}.

\begin{wraptable}{r}{0.60\textwidth}
\centering
\scriptsize
\setlength{\tabcolsep}{2pt}
\caption{Cross-family extension of Table~\ref{tab:controller-ablations}(a). Gain $=$ Acc.\ $-$ Pass@1; cell shading on the Gain column is proportional to gain magnitude (darker $=$ larger gain).}
\label{tab:cross-family}
\begin{tabularx}{\linewidth}{X X r r r}
\toprule
Orchestrator & Benchmark & P@1 & Acc. & Gain \\
\midrule
\groupheader{5}{Cross-family (orchestrator only)}
Qwen3.6-35B   & HLE-Verified  & 48.00 & 56.00 & \cellcolor{ClaudeOrange!47}+8.00 \\
Qwen3.6-35B   & LiveCodeBench & 77.14 & 78.29 & \cellcolor{ClaudeOrange!15}+1.15 \\
Gemma-4-26B   & HLE-Verified  & 48.00 & 51.00 & \cellcolor{ClaudeOrange!26}+3.00 \\
Gemma-4-26B   & LiveCodeBench & 77.14 & 80.00 & \cellcolor{ClaudeOrange!23}+2.86 \\
Gemma-4-26B   & BabyVision    & 19.59 & 21.91 & \cellcolor{ClaudeOrange!21}+2.32 \\
\groupheader{5}{Cross-family (full stack)}
Claude Sonnet & HLE-Verified  & 48.00 & 56.00 & \cellcolor{ClaudeOrange!47}+8.0 \\
GPT-5.2 (low) & HLE-Verified  & 55.00 & 56.00 & \cellcolor{ClaudeOrange!14}+1.0 \\
GPT-5.2 (high)& HLE-Verified  & 55.00 & 57.00 & \cellcolor{ClaudeOrange!19}+2.0 \\
Qwen3.6-35B   & HLE-Verified  & 19.00 & 24.00 & \cellcolor{ClaudeOrange!33}+5.0 \\
Qwen3.6-35B   & LiveCodeBench & 47.43 & 59.43 & \cellcolor{ClaudeOrange!66}+12.0 \\
Qwen3.6-35B   & BabyVision    & 16.75 & 22.94 & \cellcolor{ClaudeOrange!39}+6.2 \\
\bottomrule
\end{tabularx}
\end{wraptable}

In the orchestrator-only swap, both open-weights orchestrators deliver strictly positive Gain on every benchmark they cover: Qwen improves HLE-Verified and LiveCodeBench by $+8.00$ and $+1.15$ points, while Gemma improves HLE-Verified, LiveCodeBench, and BabyVision by $+3.00$, $+2.86$, and $+2.32$ points. In the full-stack swap, GPT-5.2 gives positive gains on HLE-Verified at both reasoning-effort levels ($+1.0$ at low effort and $+2.0$ at high effort).

The Qwen full-stack configuration acts as both explorer and orchestrator, so its Pass@1 column reflects Qwen's own first-shot accuracy (lower than Sonnet's on HLE-Verified, $19.00$ vs.\ $48.00$, and LiveCodeBench, $47.43$ vs.\ $77.14$, and on BabyVision, $16.75$ vs.\ $19.59$). Even from this lower starting point, the Qwen-driven framework lifts Acc.\ on all three benchmarks shown, with the largest absolute gain on LiveCodeBench ($+12.0$). Positive Gain in every row indicates that the explore-or-stop loop is not specific to the Anthropic family; the orchestrator only needs the in-context capability to read candidate traces and decide.

\refparagraph{sec:analysis}{Cost-accuracy balance.}
\begin{wraptable}{r}{0.30\textwidth}
\vspace{-1.0em}
\centering
\caption{Seed stability on GPQA-Diamond ($n{=}5$).}
\label{tab:seed-stability}
\small
\setlength{\tabcolsep}{4pt}
\begin{tabular}{l r r}
\toprule
            & Mean  & Std  \\
\midrule
Acc.\ (\%)  & 85.86 & 0.55 \\
\$/q        & 0.35  & 0.01 \\
\bottomrule 
\end{tabular}
\vspace{-3.0em}
\end{wraptable}
Figure~\ref{fig:cost-accuracy} gives a compact cost--accuracy summary for the Sonnet 4.6 comparison set, averaging reported accuracy and API cost over GPQA-Diamond and BabyVision. Top to bottom, each method exposes more decision dimensions to the orchestrator: LLM Selection has no exploration loop; \our{} adds adaptive stopping; \attsmm{} and \attsmi{} add model dispatch and targeted-instruction dispatch respectively. \our{} strictly dominates Majority Voting, Self-Refine without early stopping, and Budget Forcing on this Pareto frontier: higher mean accuracy at substantially lower per-question cost. LLM Selection has the highest mean cost without a corresponding accuracy gain. The multi-model variant \attsmm{} extends the same Pareto frontier further. The GPQA-Diamond stability check in Table~\ref{tab:seed-stability} gives mean accuracy $85.86\%$ with standard deviation $0.55$pp (approximately one question across runs), confirming that the \our{} point is not driven by seed noise on that component. The Pareto position is achieved by stopping early when evidence already converges; the next paragraphs trace this stopping behavior at the trajectory level.

\begin{figure}[H]
\centering
\begin{minipage}[t]{0.40\textwidth}
\centering
\vspace{0pt}
\includegraphics[width=\linewidth]{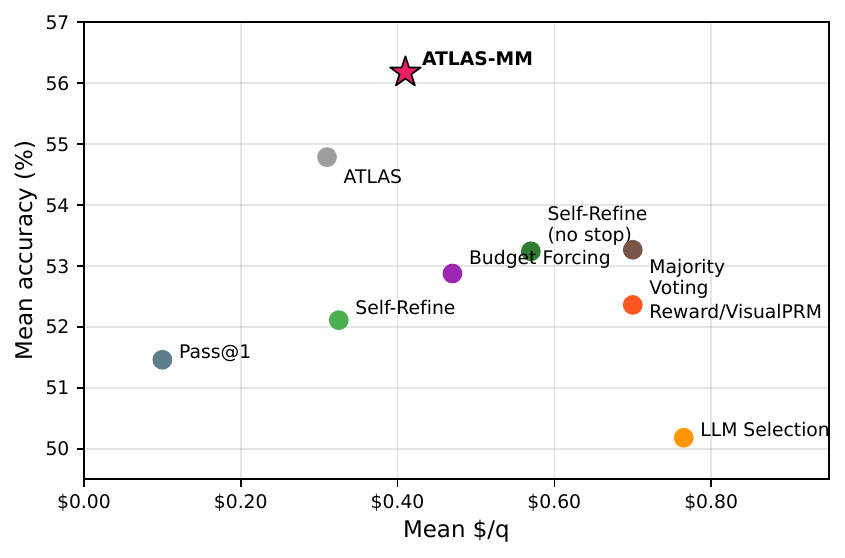}
\captionof{figure}{Cost--accuracy tradeoff summary.}
\label{fig:cost-accuracy}
\end{minipage}\hfill
\begin{minipage}[t]{0.59\textwidth}
\centering
\vspace{0pt}
\captionof{table}{Accuracy (\%) and average cost per question (\$/q) along the action-space expressiveness axis of Figure~\ref{fig:atlas-action-space}.}
\label{tab:explore-augmentation}
\scriptsize
\setlength{\tabcolsep}{1.5pt}
\begin{tabular}{l r r r r r r r r}
\toprule
 & \multicolumn{2}{c}{HLE-Verified} & \multicolumn{2}{c}{LiveCodeBench} & \multicolumn{2}{c}{BabyVision} & \multicolumn{2}{c}{GPQA-Diamond} \\
\cmidrule(lr){2-3}\cmidrule(lr){4-5}\cmidrule(lr){6-7}\cmidrule(lr){8-9}
 & Acc.$\uparrow$ & \$/q$\downarrow$ & Acc.$\uparrow$ & \$/q$\downarrow$ & Acc.$\uparrow$ & \$/q$\downarrow$ & Acc.$\uparrow$ & \$/q$\downarrow$ \\
\midrule
LLM Sel ($N{=}8$) & 58.00 & 3.71 & 81.14 & 1.51 & 23.20 & 0.56 & 77.16 & 0.97 \\
\our{}            & 56.00 & 1.59 & 82.29 & 0.53 & 23.71 & 0.27 & 85.86 & 0.35 \\
\attsmm{}         & 60.00 & 1.54 & 85.63 & 0.69 & 23.97 & 0.43 & 88.38 & 0.39 \\
\attsmi{}         & 60.00   & 1.36  & 89.14   & 0.52  & --   & --  & --   & --  \\
\bottomrule
\end{tabular}
\end{minipage}
\end{figure}

\begin{figure}[t]
  \centering
  \includegraphics[width=\textwidth]{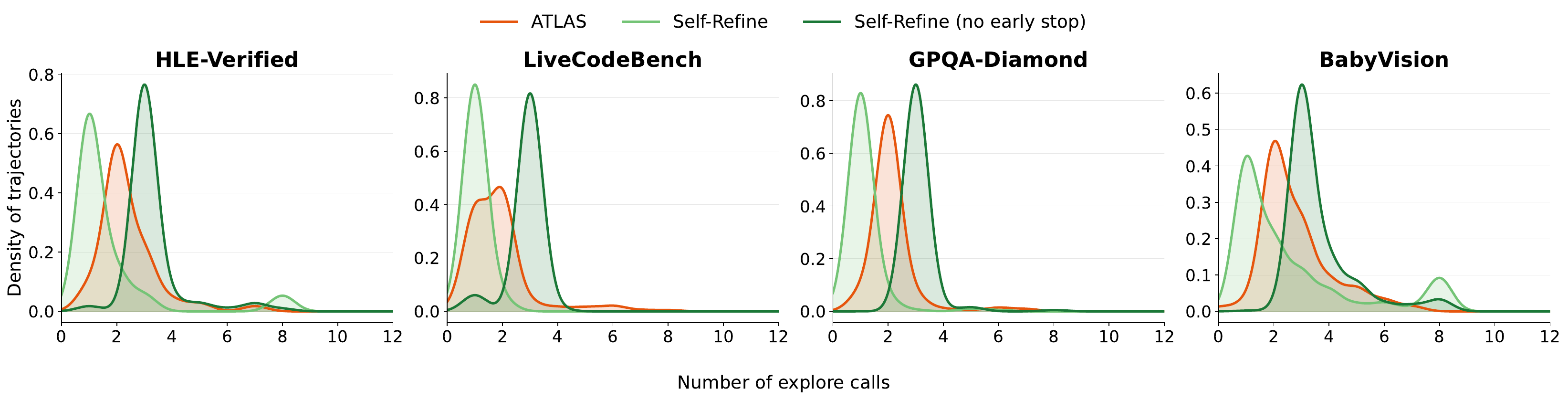}
  \caption{Per-question explore-call distributions for \our{} versus the two Self-Refine variants on the shared Sonnet 4.6 backbone, one panel per benchmark. Curves are Gaussian KDEs. Per-cell ($n$, $\mu$) and underlying discrete histograms appear in Appendix~\ref{app:explore-distributions-method}.}
  \label{fig:kde-method-comparison}
\end{figure}

To see how the cost balance above is achieved at the trajectory level, Figure~\ref{fig:kde-method-comparison} contrasts \our{}'s explore-call distribution with the two Self-Refine variants. Rather than putting most weight at $n{=}1$ (like standard Self-Refine) or always running to a fixed budget (like Self-Refine without early stopping), \our{} shows a dynamic, problem-dependent distribution with a meaningful tail. This shape confirms that the orchestrator actively adjusts compute based on evidence convergence: easy questions are resolved quickly, while harder questions receive additional exploration.

\paragraph{Ablation on the explore action space.}
Table~\ref{tab:explore-augmentation} reports accuracy and cost along the expressiveness axis defined in Figure~\ref{fig:atlas-action-space}. LLM Selection forces a fixed eight-explore budget regardless of pool state, while \our{} stops adaptively at substantially lower per-question cost on every benchmark, isolating the value of the stopping decision alone. \attsmm{} (which exposes solver choice) and \attsmi{} (which exposes a targeted-instruction argument) further improve accuracy over \our{} on HLE-Verified and LiveCodeBench, with \attsmi{} reaching the highest LiveCodeBench accuracy in the table. The transition from LLM Selection through \our{} to its augmented variants traces increasing argument expressiveness on the same loop, isolating the value of each added decision dimension.

\begin{figure}[H]
\centering
\begin{subfigure}[t]{0.49\linewidth}
  \centering
  \includegraphics[width=\linewidth]{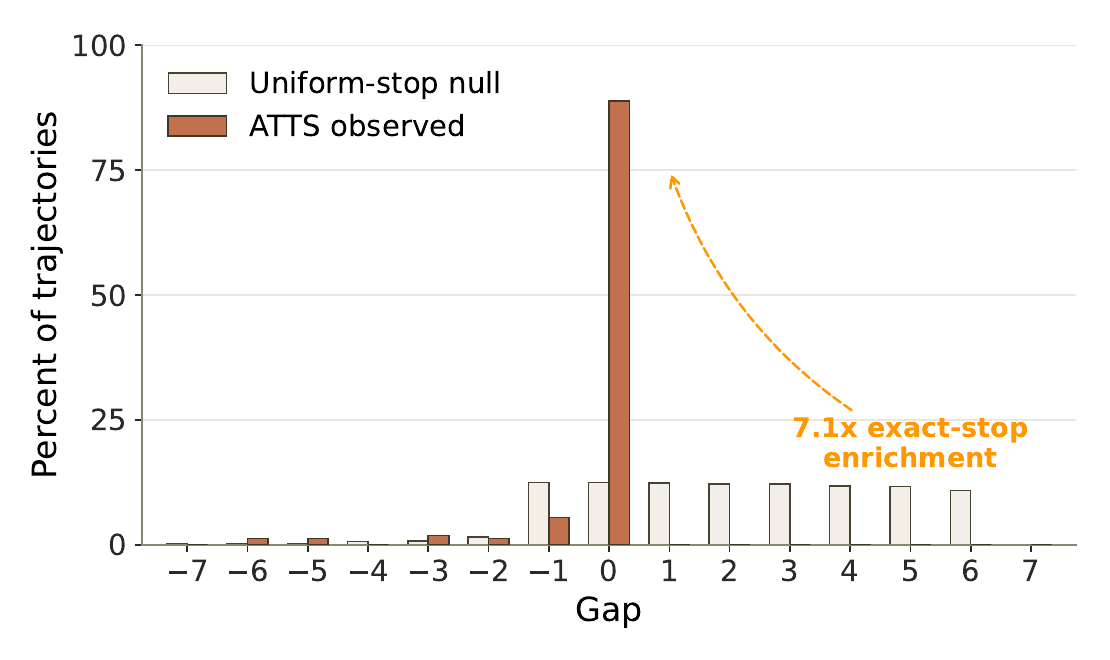}
  \caption{Stop-time gap on the $n{=}790$ GPQA-Diamond trajectories with a defined convergence point.}
  \label{fig:analysis-gap-histogram}
\end{subfigure}
\hfill
\begin{subfigure}[t]{0.49\linewidth}
  \centering
  \includegraphics[width=\linewidth]{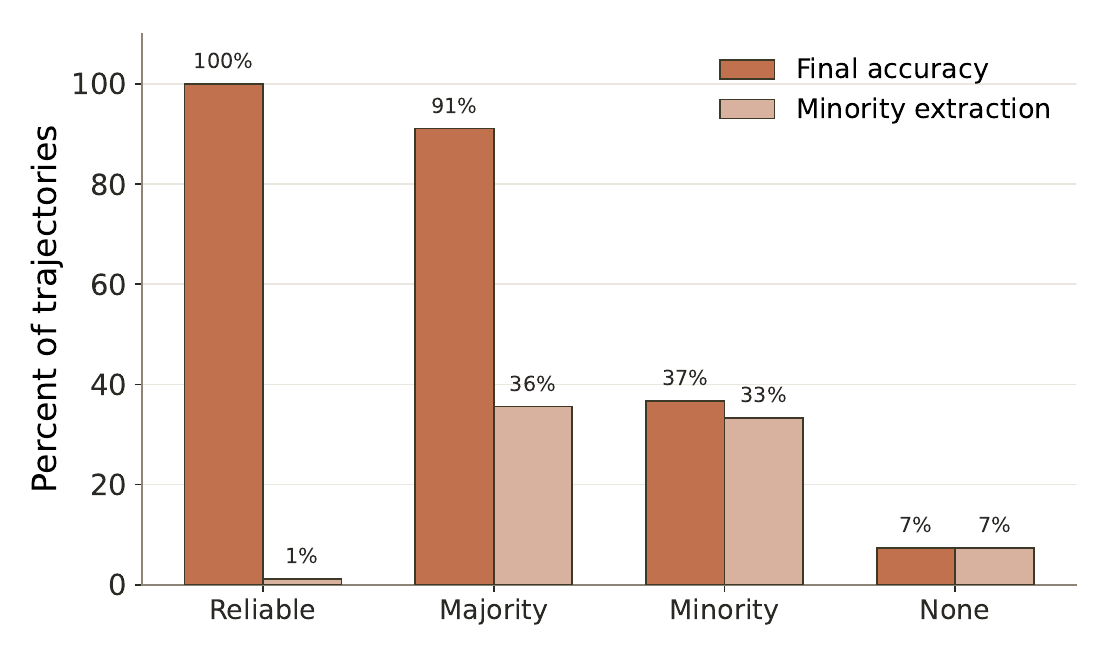}
  \caption{GPQA-Diamond behavior stratified by the number of correct candidates among eight explores.}
  \label{fig:analysis-gpqa-cohorts}
\end{subfigure}
\caption{Candidate-pool diagnostics on GPQA-Diamond. Left: the orchestrator's natural stop time aligns with correct-majority emergence rather than a random schedule. Right: performance is near-perfect when the explores contain many correct candidates and degrades as correct evidence becomes sparse.}
\label{fig:analysis-stop-time}
\end{figure}

\paragraph{Stopping and synthesis behavior.}
Figure~\ref{fig:analysis-stop-time} examines two mechanisms on GPQA-Diamond. Panel (a) compares the orchestrator's natural stop time with the earliest point at which the candidates produced so far form a correct majority; the resulting gap is sharply concentrated at zero, with $88.9\%$ of trajectories stopping exactly at convergence. Panel (b) groups the same trajectories by the number of correct candidates among the eight explores: \our{} achieves near-perfect accuracy when most candidates are correct, and its ability to recover correct minority options (reaching $36.7\%$ accuracy even when the correct option is not a majority) distinguishes it from rigid selection rules. Together, these patterns indicate that the orchestrator stops at the right moment and synthesizes across reasoning traces.

\section{Conclusion}

We introduced \our{}, an agentic test-time scaling framework in which an LLM orchestrator owns the control loop end-to-end through a single \explore{} action. Across four benchmarks, \our{} achieves competitive accuracy at substantially lower cost, and the multi-model extension \attsmm{} adds further gains by exposing solver choice as an action dimension. Ablations point to stateful evidence management as a key driver of these gains. More broadly, this shifts the design surface for test-time scaling from the controller logic to the \emph{action space}: any decision the agent should make in context can be exposed as an argument to \explore{} without retraining.

Two natural extensions remain. Explorers currently return reasoning summaries rather than full traces; richer summary contracts that pass full traces or structured summaries are a natural follow-up. Learning an orchestration policy from trajectories is a complementary direction we leave to future work.

\bibliographystyle{plainnat}
\bibliography{main}

@book{russell1991right,
  title     = {Do the Right Thing: Studies in Limited Rationality},
  author    = {Russell, Stuart J. and Wefald, Eric},
  year      = {1991},
  publisher = {MIT Press},
  address   = {Cambridge, MA},
}

@article{hle_paper_nature,
  title   = {A benchmark of expert-level academic questions to assess {AI} capabilities},
  author  = {{Center for AI Safety} and {Scale AI} and {HLE Contributors Consortium}},
  journal = {Nature},
  volume  = {649},
  pages   = {1139--1146},
  year    = {2026},
  doi     = {10.1038/s41586-025-09962-4}
}

@misc{hle_verified_paper,
  title         = {HLE-Verified: A Systematic Verification and Structured Revision of Humanity's Last Exam},
  author        = {Zhai, Weiqi and Wang, Zhihai and Wang, Jinghang and others},
  year          = {2026},
  eprint        = {2602.13964},
  archiveprefix = {arXiv},
  primaryclass  = {cs.CL},
  url           = {https://arxiv.org/abs/2602.13964}
}

@misc{hle_verified_dataset_hf,
  title        = {skylenage/HLE-Verified (Hugging Face dataset card)},
  author       = {skylenage},
  year         = {2026},
  howpublished = {\url{https://huggingface.co/datasets/skylenage/HLE-Verified}},
}

@inproceedings{livecodebench_paper,
  title     = {LiveCodeBench: Holistic and Contamination Free Evaluation of Large Language Models for Code},
  author    = {Jain, Naman and Han, Kevin and Gu, Alex and others},
  booktitle = {ICLR},
  year      = {2025},
  eprint    = {2403.07974},
  archiveprefix = {arXiv},
  url       = {https://arxiv.org/abs/2403.07974}
}

@misc{lcb_dataset_hf,
  title        = {livecodebench/code\_generation\_lite (Hugging Face dataset card)},
  author       = {livecodebench},
  year         = {2026},
  howpublished = {\url{https://huggingface.co/datasets/livecodebench/code_generation_lite}},
}

@misc{livecodebench_leaderboard,
  title        = {LiveCodeBench Leaderboard},
  author       = {{LiveCodeBench Team}},
  year         = {2026},
  howpublished = {\url{https://livecodebench.github.io/leaderboard.html}},
}

@misc{babyvision_paper,
  title         = {BabyVision: Visual Reasoning Beyond Language},
  author        = {Chen, Liang and Xie, Weichu and Liang, Yiyan and others},
  year          = {2026},
  eprint        = {2601.06521},
  archiveprefix = {arXiv},
  primaryclass  = {cs.CV},
  url           = {https://arxiv.org/abs/2601.06521}
}

@misc{babyvision_dataset_hf,
  title        = {UnipatAI/BabyVision (Hugging Face dataset card)},
  author       = {UniPat-AI},
  year         = {2026},
  howpublished = {\url{https://huggingface.co/datasets/UnipatAI/BabyVision}},
}

@misc{rbenchv_paper,
  title         = {RBench-V: A Primary Assessment for Visual Reasoning Models with Multi-modal Outputs},
  author        = {Guo, Meng-Hao and Chu, Xuanyu and Yang, Qianrui and others},
  year          = {2025},
  eprint        = {2505.16770},
  archiveprefix = {arXiv},
  primaryclass  = {cs.CV},
  url           = {https://arxiv.org/abs/2505.16770}
}

@misc{rbenchv_dataset_hf,
  title        = {R-Bench/R-Bench-V (Hugging Face dataset card and leaderboard)},
  author       = {{R-Bench Team}},
  year         = {2026},
  howpublished = {\url{https://huggingface.co/datasets/R-Bench/R-Bench-V}},
}

@inproceedings{gpqa_paper,
  title     = {GPQA: A Graduate-Level Google-Proof Q\&A Benchmark},
  author    = {Rein, David and Hou, Betty Li and Stickland, Asa Cooper and others},
  booktitle = {First Conference on Language Modeling (COLM)},
  year      = {2024},
  eprint    = {2311.12022},
  archiveprefix = {arXiv},
  url       = {https://arxiv.org/abs/2311.12022}
}

@misc{gpqa_dataset_hf,
  title        = {Idavidrein/gpqa (Hugging Face dataset card)},
  author       = {Idavidrein},
  year         = {2026},
  howpublished = {\url{https://huggingface.co/datasets/Idavidrein/gpqa}},
}

@inproceedings{snell_test_time_scaling_paper,
  title     = {Scaling {LLM} Test-Time Compute Optimally can be More Effective than Scaling Model Parameters},
  author    = {Snell, Charlie and Lee, Jaehoon and Xu, Kelvin and Kumar, Aviral},
  booktitle = {International Conference on Learning Representations},
  year      = {2025},
  url       = {https://arxiv.org/abs/2408.03314}
}

@inproceedings{self_consistency_paper,
  title     = {Self-Consistency Improves Chain of Thought Reasoning in Language Models},
  author    = {Wang, Xuezhi and Wei, Jason and Schuurmans, Dale and Le, Quoc and Chi, Ed and Narang, Sharan and Chowdhery, Aakanksha and Zhou, Denny},
  booktitle = {International Conference on Learning Representations},
  year      = {2023},
  url       = {https://arxiv.org/abs/2203.11171}
}

@inproceedings{self_refine_paper,
  title     = {Self-Refine: Iterative Refinement with Self-Feedback},
  author    = {Madaan, Aman and Tandon, Niket and Gupta, Prakhar and Hallinan, Skyler and Gao, Luyu and Wiegreffe, Sarah and Alon, Uri and Dziri, Nouha and Prabhumoye, Shrimai and Yang, Yiming and Welleck, Sean and Majumder, Bodhisattwa Prasad and Gupta, Shashank and Yazdanbakhsh, Amir and Clark, Peter},
  booktitle = {Advances in Neural Information Processing Systems},
  year      = {2023},
  url       = {https://proceedings.neurips.cc/paper_files/paper/2023/file/91edff07232fb1b55a505a9e9f6c0ff3-Paper-Conference.pdf}
}

@inproceedings{s1_budget_forcing_paper,
  title     = {s1: Simple test-time scaling},
  author    = {Muennighoff, Niklas and Yang, Zitong and Shi, Weijia and Li, Xiang Lisa and Fei-Fei, Li and Hajishirzi, Hannaneh and Zettlemoyer, Luke and Liang, Percy and Cand{\`e}s, Emmanuel and Hashimoto, Tatsunori},
  booktitle = {Proceedings of the 2025 Conference on Empirical Methods in Natural Language Processing},
  pages     = {20275--20321},
  year      = {2025},
  publisher = {Association for Computational Linguistics},
  doi       = {10.18653/v1/2025.emnlp-main.1025},
  url       = {https://aclanthology.org/2025.emnlp-main.1025/}
}

@inproceedings{best_of_n_optimality_paper,
  title         = {Is Best-of-$N$ the Best of Them? Coverage, Scaling, and Optimality in Inference-Time Alignment},
  author        = {Huang, Audrey and Block, Adam and Liu, Qinghua and Jiang, Nan and Krishnamurthy, Akshay and Foster, Dylan J.},
  booktitle     = {International Conference on Machine Learning (ICML)},
  year          = {2025},
  url           = {https://arxiv.org/abs/2503.21878}
}

@misc{visualprm_paper,
  title         = {VisualPRM: An Effective Process Reward Model for Multimodal Reasoning},
  author        = {Wang, Weiyun and Gao, Zhangwei and Chen, Lianjie and Chen, Zhe and Zhu, Jinguo and Zhao, Xiangyu and Liu, Yangzhou and Cao, Yue and Ye, Shenglong and Zhu, Xizhou and Lu, Lewei and Duan, Haodong and Qiao, Yu and Dai, Jifeng and Wang, Wenhai},
  year          = {2025},
  eprint        = {2503.10291},
  archiveprefix = {arXiv},
  primaryclass  = {cs.CV},
  url           = {https://arxiv.org/abs/2503.10291}
}

@inproceedings{tree_of_thoughts_paper,
  title         = {Tree of Thoughts: Deliberate Problem Solving with Large Language Models},
  author        = {Yao, Shunyu and Yu, Dian and Zhao, Jeffrey and Shafran, Izhak and Griffiths, Thomas L. and Cao, Yuan and Narasimhan, Karthik},
  booktitle     = {Advances in Neural Information Processing Systems},
  year          = {2023},
  url           = {https://arxiv.org/abs/2305.10601}
}

@inproceedings{react_paper,
  title         = {{ReAct}: Synergizing Reasoning and Acting in Language Models},
  author        = {Yao, Shunyu and Zhao, Jeffrey and Yu, Dian and Du, Nan and Shafran, Izhak and Narasimhan, Karthik and Cao, Yuan},
  booktitle     = {International Conference on Learning Representations},
  year          = {2023},
  url           = {https://arxiv.org/abs/2210.03629}
}

@misc{mcts_reasoning_paper,
  title         = {Monte Carlo Tree Search Boosts Reasoning via Iterative Preference Learning},
  author        = {Xie, Yuxi and Goyal, Anirudh and Zheng, Wenyue and Kan, Min-Yen and Lillicrap, Timothy P. and Kawaguchi, Kenji and Shieh, Michael},
  year          = {2024},
  eprint        = {2405.00451},
  archiveprefix = {arXiv},
  primaryclass  = {cs.LG},
  url           = {https://arxiv.org/abs/2405.00451}
}

@inproceedings{math_shepherd_paper,
  title         = {Math-Shepherd: Verify and Reinforce {LLM}s Step-by-step without Human Annotations},
  author        = {Wang, Peiyi and Li, Lei and Shao, Zhihong and Xu, R. X. and Dai, Damai and Li, Yifei and Chen, Deli and Wu, Yiran and Sui, Zhifang},
  booktitle     = {Proceedings of the 62nd Annual Meeting of the Association for Computational Linguistics (ACL)},
  year          = {2024},
  url           = {https://arxiv.org/abs/2312.08935}
}

@misc{rsa_paper,
  title         = {Recursive Self-Aggregation Unlocks Deep Thinking in Large Language Models},
  author        = {Venkatraman, Siddarth and Jain, Vineet and Mittal, Sarthak and Shah, Vedant and Obando-Ceron, Johan and Bengio, Yoshua and Bartoldson, Brian R. and Kailkhura, Bhavya and Lajoie, Guillaume and Berseth, Glen and Malkin, Nikolay and Jain, Moksh},
  year          = {2025},
  eprint        = {2509.26626},
  archiveprefix = {arXiv},
  primaryclass  = {cs.LG},
  url           = {https://arxiv.org/abs/2509.26626}
}

@inproceedings{skywork_reward_v2_paper,
  title         = {Skywork-Reward-V2: Scaling Preference Data Curation via Human-AI Synergy},
  author        = {Liu, Chris Yuhao and Zeng, Liang and Xiao, Yuzhen and He, Jujie and Liu, Jiacai and Wang, Chaojie and Yan, Rui and Shen, Wei and Zhang, Fuxiang and Xu, Jiacheng and Liu, Yang and Zhou, Yahui},
  booktitle     = {International Conference on Learning Representations},
  year          = {2026},
  eprint        = {2507.01352},
  archiveprefix = {arXiv},
  url           = {https://arxiv.org/abs/2507.01352}
}

@misc{catts_webagents_paper,
  title         = {Agentic Test-Time Scaling for WebAgents},
  author        = {Lee, Nicholas and Erdogan, Lutfi Eren and John, Chris Joseph and Krishnapillai, Surya and Mahoney, Michael W. and Keutzer, Kurt and Gholami, Amir},
  year          = {2026},
  eprint        = {2602.12276},
  archiveprefix = {arXiv},
  primaryclass  = {cs.AI},
  url           = {https://arxiv.org/abs/2602.12276}
}

@misc{general_agentbench_tts_paper,
  title         = {Benchmark Test-Time Scaling of General {LLM} Agents},
  author        = {Li, Xiaochuan and Ming, Ryan and Setlur, Pranav and Paladugu, Abhijay and Tang, Andy and Kang, Hao and Shao, Shuai and Jin, Rong and Xiong, Chenyan},
  year          = {2026},
  eprint        = {2602.18998},
  archiveprefix = {arXiv},
  primaryclass  = {cs.AI},
  url           = {https://arxiv.org/abs/2602.18998}
}

@misc{tumix_paper,
  title         = {TUMIX: Multi-Agent Test-Time Scaling with Tool-Use Mixture},
  author        = {Chen, Yongchao and Chen, Jiefeng and Meng, Rui and Yin, Ji and Li, Na and Fan, Chuchu and Wang, Chi and Pfister, Tomas and Yoon, Jinsung},
  year          = {2025},
  eprint        = {2510.01279},
  archiveprefix = {arXiv},
  primaryclass  = {cs.CL},
  url           = {https://arxiv.org/abs/2510.01279}
}

@misc{skywork_reward_v2_hf,
  title        = {Skywork-Reward-V2-Qwen3-8B (Hugging Face model card)},
  author       = {Skywork},
  year         = {2025},
  howpublished = {\url{https://huggingface.co/Skywork/Skywork-Reward-V2-Qwen3-8B}},
}

@misc{visualprm_hf,
  title        = {VisualPRM-8B (Hugging Face model card)},
  author       = {OpenGVLab},
  year         = {2025},
  howpublished = {\url{https://huggingface.co/OpenGVLab/VisualPRM-8B}},
}

@misc{claude_agent_sdk,
  title        = {Claude Code {SDK} Documentation},
  author       = {Anthropic},
  year         = {2025},
  howpublished = {\url{https://docs.anthropic.com/en/docs/claude-code/sdk}},
}

@misc{large_language_monkeys_paper,
  title         = {Large Language Monkeys: Scaling Inference Compute with Repeated Sampling},
  author        = {Brown, Bradley and Juravsky, Jordan and Ehrlich, Ryan and Clark, Ronald and Le, Quoc V. and R{\'e}, Christopher and Mirhoseini, Azalia},
  year          = {2024},
  eprint        = {2407.21787},
  archiveprefix = {arXiv},
  primaryclass  = {cs.CL},
  url           = {https://arxiv.org/abs/2407.21787}
}

@inproceedings{universal_self_consistency_paper,
  title         = {Universal Self-Consistency for Large Language Model Generation},
  author        = {Chen, Xinyun and Aksitov, Renat and Alon, Uri and Ren, Jie and Xiao, Kefan and Yin, Pengcheng and Prakash, Sushant and Sutton, Charles and Wang, Xuezhi and Zhou, Denny},
  booktitle     = {ICML 2024 Workshop on In-Context Learning},
  year          = {2024},
  eprint        = {2311.17311},
  archiveprefix = {arXiv},
  url           = {https://arxiv.org/abs/2311.17311}
}

@inproceedings{adaptive_inference_time_paper,
  title         = {Adaptive Inference-Time Compute: {LLMs} Can Predict if They Can Do Better, Even Mid-Generation},
  author        = {Manvi, Rohin and Singh, Anikait and Ermon, Stefano},
  booktitle     = {International Conference on Learning Representations},
  year          = {2025},
  eprint        = {2410.02725},
  archiveprefix = {arXiv},
  url           = {https://arxiv.org/abs/2410.02725}
}

@inproceedings{rational_metareasoning_llm_paper,
  title         = {Rational Metareasoning for Large Language Models},
  author        = {De Sabbata, C. Nicol{\`o} and Sumers, Theodore R. and AlKhamissi, Badr and Bosselut, Antoine and Griffiths, Thomas L.},
  booktitle     = {Advances in Neural Information Processing Systems},
  year          = {2024},
  eprint        = {2410.05563},
  archiveprefix = {arXiv},
  url           = {https://arxiv.org/abs/2410.05563}
}

@inproceedings{cannot_self_correct_paper,
  title         = {Large Language Models Cannot Self-Correct Reasoning Yet},
  author        = {Huang, Jie and Chen, Xinyun and Mishra, Swaroop and Zheng, Huaixiu Steven and Yu, Adams Wei and Song, Xinying and Zhou, Denny},
  booktitle     = {International Conference on Learning Representations},
  year          = {2024},
  eprint        = {2310.01798},
  archiveprefix = {arXiv},
  url           = {https://arxiv.org/abs/2310.01798}
}

@article{deepseek_r1_paper,
  title         = {{DeepSeek-R1} incentivizes reasoning in {LLMs} through reinforcement learning},
  author        = {{DeepSeek-AI}},
  journal       = {Nature},
  volume        = {645},
  pages         = {633--638},
  year          = {2025},
  doi           = {10.1038/s41586-025-09422-z},
  eprint        = {2501.12948},
  archiveprefix = {arXiv},
  url           = {https://www.nature.com/articles/s41586-025-09422-z}
}

@inproceedings{debate_or_vote_paper,
  title         = {Debate or Vote: Which Yields Better Decisions in Multi-Agent Large Language Models?},
  author        = {Choi, Hyeong Kyu and Zhu, Banghua},
  booktitle     = {Advances in Neural Information Processing Systems},
  year          = {2025},
  eprint        = {2508.17536},
  archiveprefix = {arXiv},
  url           = {https://arxiv.org/abs/2508.17536}
}

@inproceedings{damani_adaptive_allocation_paper,
  title         = {Learning How Hard to Think: Input-Adaptive Allocation of {LM} Computation},
  author        = {Damani, Mehul and Shenfeld, Idan and Peng, Andi and Bobu, Andreea and Andreas, Jacob},
  booktitle     = {International Conference on Learning Representations},
  year          = {2025},
  url           = {https://arxiv.org/abs/2410.04707}
}

@article{wan2025beacon,
  title   = {{BEACON}: Bayesian Optimal Stopping for Efficient {LLM} Sampling},
  author  = {Wan, Guangya and Xu, Zixin Stephen and Zorc, Sasa and Baucells, Manel and Hu, Mengxuan and Wang, Hao and Li, Sheng},
  journal = {arXiv preprint arXiv:2510.15945},
  year    = {2025},
  url     = {https://arxiv.org/abs/2510.15945}
}

@inproceedings{ab_mcts_paper,
  title         = {Wider or Deeper? Scaling {LLM} Inference-Time Compute with Adaptive Branching Tree Search},
  author        = {Inoue, Yuichi and Misaki, Kou and Imajuku, Yuki and Kuroki, So and Nakamura, Taishi and Akiba, Takuya},
  booktitle     = {Advances in Neural Information Processing Systems},
  year          = {2025},
  url           = {https://arxiv.org/abs/2503.04412}
}

@inproceedings{hay_selecting_computations,
  title         = {Selecting Computations: Theory and Applications},
  author        = {Hay, Nicholas and Russell, Stuart and Tolpin, David and Shimony, Solomon Eyal},
  year          = {2012},
  booktitle     = {Uncertainty in Artificial Intelligence},
  pages         = {346--355},
}

@article{kalayci_optimal_stopping_bon,
  title={Optimal Stopping vs Best-of-$ N $ for Inference Time Optimization},
  author={Kalayci, Yusuf and Raman, Vinod and Dughmi, Shaddin},
  journal={arXiv preprint arXiv:2510.01394},
  year={2025}
}

@inproceedings{router_r1_paper,
  title         = {Router-R1: Teaching {LLMs} Multi-Round Routing and Aggregation via Reinforcement Learning},
  author        = {Zhang, Haozhen and Feng, Tao and You, Jiaxuan},
  booktitle     = {Advances in Neural Information Processing Systems},
  year          = {2025},
  eprint        = {2506.09033},
  archiveprefix = {arXiv},
  url           = {https://arxiv.org/abs/2506.09033}
}

@article{xrouter_paper,
  title={xRouter: Training Cost-Aware LLMs Orchestration System via Reinforcement Learning},
  author={Qian, Cheng and Liu, Zuxin and Kokane, Shirley and Prabhakar, Akshara and Qiu, Jielin and Chen, Haolin and Liu, Zhiwei and Ji, Heng and Yao, Weiran and Heinecke, Shelby and others},
  journal={arXiv preprint arXiv:2510.08439},
  year={2025}
}

@inproceedings{best_route_paper,
  title={BEST-Route: Adaptive LLM Routing with Test-Time Optimal Compute},
  author={Ding, Dujian and Mallick, Ankur and Zhang, Shaokun and Wang, Chi and Madrigal, Daniel and Garcia, Mirian Del Carmen Hipolito and Xia, Menglin and Lakshmanan, Laks VS and Wu, Qingyun and R{\"u}hle, Victor},
  booktitle={International Conference on Machine Learning},
  pages={13870--13884},
  year={2025},
  organization={PMLR}
}

@inproceedings{adaptive_consistency_paper,
  title         = {Let's Sample Step by Step: Adaptive-Consistency for Efficient Reasoning and Coding with {LLMs}},
  author        = {Aggarwal, Pranjal and Madaan, Aman and Yang, Yiming and Mausam},
  booktitle     = {Proceedings of the 2023 Conference on Empirical Methods in Natural Language Processing},
  pages         = {12375--12396},
  year          = {2023},
  publisher     = {Association for Computational Linguistics},
  eprint        = {2305.11860},
  archiveprefix = {arXiv},
  url           = {https://aclanthology.org/2023.emnlp-main.761/}
}

@misc{consol_sprt_paper,
  title         = {{ConSol}: Sequential Probability Ratio Testing to Find Consistent {LLM} Reasoning Paths Efficiently},
  author        = {Lee, Jaeyeon and Qi, Guantong and Neeley, Matthew Brady and Liu, Zhandong and Jeong, Hyun-Hwan},
  year          = {2025},
  eprint        = {2503.17587},
  archiveprefix = {arXiv},
  url           = {https://arxiv.org/abs/2503.17587}
}

@inproceedings{search_r1_paper,
  title={Search-R1: Training LLMs to Reason and Leverage Search Engines with Reinforcement Learning},
  author={Jin, Bowen and Zeng, Hansi and Yue, Zhenrui and Yoon, Jinsung and Arik, Sercan O and Wang, Dong and Zamani, Hamed and Han, Jiawei},
  booktitle={Second Conference on Language Modeling},
  year={2025}
}

@article{optimal_bayesian_stopping_paper,
  title={Optimal Bayesian Stopping for Efficient Inference of Consistent LLM Answers},
  author={Huang, Jingkai and Ma, Will and Zhou, Zhengyuan},
  journal={arXiv preprint arXiv:2602.05395},
  year={2026}
}

@article{certified_self_consistency_paper,
  title={Certified self-consistency: Statistical guarantees and test-time training for reliable reasoning in LLMs},
  author={Cordero-Encinar, Paula and Duncan, Andrew B},
  journal={arXiv preprint arXiv:2510.17472},
  year={2025}
}

@inproceedings{catp_llm_paper,
  title         = {{CATP-LLM}: Empowering Large Language Models for Cost-Aware Tool Planning},
  author        = {Wu, Duo and Wang, Jinghe and Meng, Yuan and Zhang, Yanning and Sun, Le and Wang, Zhi},
  booktitle     = {Proceedings of the IEEE/CVF International Conference on Computer Vision (ICCV)},
  pages         = {8699--8709},
  year          = {2025},
  eprint        = {2411.16313},
  archiveprefix = {arXiv},
  url           = {https://arxiv.org/abs/2411.16313}
}

@misc{calibrate_then_act_paper,
  title         = {Calibrate-Then-Act: Cost-Aware Exploration in {LLM} Agents},
  author        = {Ding, Wenxuan and Tomlin, Nicholas and Durrett, Greg},
  year          = {2026},
  url           = {https://arxiv.org/abs/2602.16699}
}

@inproceedings{toolrl_paper,
  title         = {{ToolRL}: Reward is All Tool Learning Needs},
  author        = {Qian, Cheng and Acikgoz, Emre Can and He, Qi and Wang, Hongru and Chen, Xiusi and Hakkani-Tur, Dilek and Tur, Gokhan and Ji, Heng},
  booktitle     = {Advances in Neural Information Processing Systems},
  year          = {2025},
  eprint        = {2504.13958},
  archiveprefix = {arXiv},
  url           = {https://arxiv.org/abs/2504.13958}
}

@inproceedings{steptool_paper,
  title         = {{StepTool}: Enhancing Multi-Step Tool Usage in {LLMs} via Step-Grained Reinforcement Learning},
  author        = {Yu, Yuanqing and Wang, Zhefan and Ma, Weizhi and Wang, Shuai and Wu, Chuhan and Guo, Zhiqiang and Zhang, Min},
  booktitle     = {Proceedings of the 34th ACM International Conference on Information and Knowledge Management},
  year          = {2025},
  doi           = {10.1145/3746252.3761391},
  eprint        = {2410.07745},
  archiveprefix = {arXiv},
  url           = {https://dl.acm.org/doi/10.1145/3746252.3761391}
}

@inproceedings{
tool_call_rm_paper,
title={Empowering {LLM} Tool Invocation with Tool-call Reward Model},
author={Da Ma and Ziyue Yang and Hongshen Xu and Haotian Fang and Kai Yu and Lu Chen},
booktitle={The Fourteenth International Conference on Learning Representations},
year={2026},
url={https://openreview.net/forum?id=LnBEASInVr}
}

@misc{msv_paper,
  title         = {Parallel Test-Time Scaling with Multi-Sequence Verifiers},
  author        = {Kim, Yegon and Lee, Seungyoo and Jang, Chaeyun and Lee, Hyungi and Lee, Juho},
  year          = {2026},
  url           = {https://arxiv.org/abs/2603.03417}
}

@misc{corefine_paper,
  title         = {{CoRefine}: Confidence-Guided Self-Refinement for Adaptive Test-Time Compute},
  author        = {Jin, Chen and Tanno, Ryutaro and Diethe, Tom and Teare, Philip},
  year          = {2026},
  url           = {https://arxiv.org/abs/2602.08948}
}

@misc{meta_reasoner_paper,
  title         = {Meta-Reasoner: Dynamic Guidance for Optimized Inference-time Reasoning in Large Language Models},
  author        = {Sui, Yuan and He, Yufei and Cao, Tri and Han, Simeng and Chen, Yulin and Hooi, Bryan},
  year          = {2025},
  url           = {https://arxiv.org/abs/2502.19918}
}

@article{orch_paper,
  title         = {{ORCH}: many analyses, one merge---a deterministic multi-agent orchestrator for discrete-choice reasoning with {EMA}-guided routing},
  author        = {Zhou, Hanlin and Chan, Huah Yong},
  journal       = {Frontiers in Artificial Intelligence},
  volume        = {9},
  year          = {2026},
  doi           = {10.3389/frai.2026.1748735},
  url           = {https://www.frontiersin.org/journals/artificial-intelligence/articles/10.3389/frai.2026.1748735/full}
}

@article{qwen3_paper,
  title={Qwen3 technical report},
  author={Yang, An and Li, Anfeng and Yang, Baosong and Zhang, Beichen and Hui, Binyuan and Zheng, Bo and Yu, Bowen and Gao, Chang and Huang, Chengen and Lv, Chenxu and others},
  journal={arXiv preprint arXiv:2505.09388},
  year={2025}
}

\appendix

\section{Extended Related Work}
\label{app:related-work-extended}

This appendix preserves the full version of the related-work discussion summarized in Section~\ref{sec:related-work-conceptually}.

\paragraph{Test-Time Scaling.} Inference-time compute for LLMs is commonly organized along three axes. \emph{Sequential scaling} increases reasoning depth within a single trajectory, including self-correction \citep{self_refine_paper}, budget-controlled decoding \citep{s1_budget_forcing_paper}, and long-thinking models that internalize extended chains of thought via reinforcement learning \citep{deepseek_r1_paper}. A critical negative result is that LLMs cannot self-correct reasoning through prompting alone \citep{cannot_self_correct_paper}, motivating architectures that separate evaluation from solution generation. \emph{Parallel scaling} increases reasoning width by sampling multiple independent candidates and aggregating them. Self-consistency marginalizes over diverse reasoning paths \citep{self_consistency_paper}; coverage scales log-linearly with the sample count \citep{large_language_monkeys_paper}; and Universal Self-Consistency uses the LLM itself to select among candidates for free-form tasks \citep{universal_self_consistency_paper}. Learned reward models and verifiers improve selection beyond majority voting \citep{snell_test_time_scaling_paper,skywork_reward_v2_paper,visualprm_paper}, and Multi-Sequence Verifiers jointly process the entire candidate pool rather than scoring each independently \citep{msv_paper}. Theoretical analysis shows that naive parallel scaling is not uniformly beneficial: overly aggressive sampling can degrade accuracy through reward hacking \citep{best_of_n_optimality_paper}. \emph{Hybrid search} combines depth and width by expanding and pruning partial trajectories via tree search \citep{tree_of_thoughts_paper}, Monte Carlo Tree Search with preference learning \citep{mcts_reasoning_paper}, step-level process reward models \citep{math_shepherd_paper}, or interleaved aggregation and refinement \citep{rsa_paper}. \our{} is neither a single-trajectory method nor a tree-search method: it delegates \emph{independent complete solves} to a separate explorer and manages the resulting candidate pool sequentially, occupying a distinct point in this design space.

\paragraph{Adaptive Compute Allocation.} A growing body of work addresses \emph{when} to stop sampling and \emph{how much} compute to allocate per problem. Statistical stopping criteria include Beta/Dirichlet confidence thresholds \citep{adaptive_consistency_paper}, the Sequential Probability Ratio Test \citep{consol_sprt_paper}, anytime-valid concentration bounds \citep{certified_self_consistency_paper}, Bayesian optimal stopping with sequential posterior updates \citep{wan2025beacon}, and optimal answer tracking showing that monitoring the top-3 answers suffices for asymptotic optimality \citep{optimal_bayesian_stopping_paper}. \citet{kalayci_optimal_stopping_bon} frame each generation as opening a box in Weitzman's Pandora's Box problem, deriving reservation-price stopping rules. On the learned side, \citet{damani_adaptive_allocation_paper} train a difficulty predictor to adapt the number of best-of-$k$ samples per query, and BEST-Route jointly selects which model and how many responses to sample \citep{best_route_paper}. These methods rely on either statistical criteria or learned scalar predictors. At the theoretical level, \citet{rational_metareasoning_llm_paper} operationalize rational metareasoning for LLMs at the \emph{token level}, training models with a VOC-derived reward that penalizes unnecessary reasoning steps within a single generation; they identify candidate-level agentic extension as an open direction. \our{} uses an LLM orchestrator that reads full reasoning traces (accessing richer information than answer frequencies or scalar scores) within a sequential closed loop where each observation informs the next action.

\paragraph{Agentic Orchestration.} Recent work studies both agentic test-time scaling and learned meta-level controllers. On the scaling side, CATTS allocates compute via vote-derived uncertainty \citep{catts_webagents_paper}, TUMIX coordinates parallel heterogeneous agents \citep{tumix_paper}, and General AgentBench benchmarks sequential and parallel scaling for open-ended agents \citep{general_agentbench_tts_paper}. On the meta-control side, \citet{adaptive_inference_time_paper} show that LLMs can predict mid-generation whether restarting will help; CoRefine attaches a lightweight controller that decides to halt, re-examine, or try a different approach \citep{corefine_paper}; Meta-Reasoner employs a contextual bandit over high-level strategies such as backtrack, restart, or switch approach \citep{meta_reasoner_paper}; AB-MCTS dynamically decides between expanding width and depth \citep{ab_mcts_paper}; and ORCH orchestrates multiple LLM analyses followed by a single merge \citep{orch_paper}. A parallel trend trains LLM orchestrators via reinforcement learning: Router-R1 uses GRPO for multi-round routing to external LLMs \citep{router_r1_paper}, xRouter learns cost-aware orchestration across multiple external models \citep{xrouter_paper}, and Search-R1 trains multi-turn search-interleaved reasoning \citep{search_r1_paper}. On multi-agent interaction, symmetric debate has been shown to induce a martingale over belief trajectories, providing no expected accuracy improvement beyond majority voting \citep{debate_or_vote_paper}. \our{} differs from these approaches in one key respect: it dispatches \emph{fresh independent solvers} (not routing to different models or managing a single evolving trajectory), and only the orchestrator accumulates evidence. The result is an asymmetric, delegated architecture with a single action type (\explore{}) whose value is assessed after each invocation.

\paragraph{Cost-Aware Agent Training.} A nascent line of work incorporates execution costs into agent objectives. CATP-LLM optimizes a performance-cost tradeoff for tool planning via cost-aware offline RL \citep{catp_llm_paper}. Calibrate-Then-Act formalizes the cost-uncertainty tradeoff for agent exploration, comparing the expected information value of an additional tool call against its cost \citep{calibrate_then_act_paper}. On reward engineering for tool use, ToolRL systematically studies reward type, scale, and granularity for GRPO-based training \citep{toolrl_paper}; StepTool assigns per-interaction rewards based on invocation success and task contribution \citep{steptool_paper}; and the Tool-call Reward Model enables step-level credit assignment in PPO/GRPO pipelines \citep{tool_call_rm_paper}. These works address planning, exploration, and reward design for general tool use. \our{} targets a more specific problem: \emph{candidate-level test-time scaling}, where the ``tool'' is an independent solver call and the meta-decision is a sequential explore-or-stop choice. To our knowledge, no prior work combines an LLM orchestrator that reads full candidate reasoning traces, sequential closed-loop stopping, and independently generated candidates within a unified framework for test-time compute allocation.

\section{Implementation Details}
\label{app:implementation-details}

\paragraph{Controller and backend setup.}
\our{} is implemented as a delegated controller with two roles: an orchestrator and an explorer. The orchestrator never solves the task directly; it only decides whether to spend the remaining budget on another \explore{} call or to stop and synthesize the final answer from the accumulated evidence. Each \explore{} call returns a structured candidate containing an answer, supporting reasoning, an approach summary, and an optional confidence score. When the orchestrator decides to stop, it produces the final answer directly via the SDK's structured output mechanism. For the Claude implementation, we use the Claude Agent SDK \citep{claude_agent_sdk} and expose \explore{} as the only available delegated tool. We also evaluate a variant that adds a separate \integrate{} tool that delegates final synthesis to a dedicated model call; see the ablation in Table~\ref{tab:controller-ablations}(b). The same high-level controller is reused across benchmarks, while the underlying sub-model backend can be swapped without changing the controller logic.

\paragraph{Variant configurations and sampling.}
The two configurations reported in the main results are concrete instantiations of the variants in \S\ref{sec:variants}. \our{} (single-model) uses Claude Sonnet 4.6 (\texttt{claude-sonnet-4-6}) for both the orchestrator and the explorer, with exploration budget $T{=}8$ and direct orchestrator synthesis (no \integrate{}). \attsmm{} (multi-model) keeps Sonnet 4.6 as the orchestrator and dispatches to a three-model pool $\{\text{Haiku}, \text{Sonnet}, \text{Opus}\}$ at \emph{Medium} explore effort, with budget $T{=}20$ split as Haiku~8, Sonnet~8, Opus~4. Across all variants and baselines, sampling parameters use the API defaults, and Claude orchestrator and explorer calls run with extended thinking enabled at the API's default reasoning effort, except where stated otherwise; the extended-thinking budget is 32{,}000 tokens per call.

\paragraph{Open-weights orchestrator backends.}
For the cross-family ablations (Table~\ref{tab:cross-family}), we use two open-weights backbones served locally via vLLM:
\begin{itemize}
\item \textbf{Qwen3.6-35B-A3B}: Hugging Face checkpoint
\texttt{Qwen/Qwen3.6-35B-A3B-Instruct-FP8} (\url{https://huggingface.co/Qwen/Qwen3.6-35B-A3B-Instruct-FP8}). A Mixture-of-Experts model with 35B total / 3B active parameters, served as native FP8 with tensor-parallel size 2. We use Qwen3.6's recommended thinking-mode decoding: $T{=}1.0$, top-$p{=}0.95$, top-$k{=}20$, presence-penalty $1.5$, with thinking mode enabled.
\item \textbf{Gemma-4-26B-A4B}: Hugging Face checkpoint \texttt{google/gemma-4-26B-A4B-it} (\url{https://huggingface.co/google/gemma-4-26B-A4B-it}). A Mixture-of-Experts model with 26B total / 4B active parameters, served via vLLM. We decode with thinking mode disabled; the orchestrator aggregates over candidates produced by a Sonnet 4.6 explorer.
\end{itemize}

\paragraph{Prompt families.}
\our{} uses two prompt families: a benchmark-agnostic orchestrator prompt and a benchmark-specific explorer prompt. The orchestrator prompt defines the controller role. The explorer prompt asks for an independent solution from scratch under the target answer format. The full templates -- orchestrator variants, stopping-threshold instructions, and finalize instructions -- are reproduced verbatim in Appendix~\ref{app:prompt-families}. For Claude runs, these prompts are paired with structured-output schemas so that each sub-call returns machine-readable fields rather than free-form text only.

\paragraph{Schemas and benchmark-specific prompt instantiation.}
Benchmark-specific adaptation is intentionally lightweight. Most answer-based tasks share the explorer schema \texttt{\{approach, reasoning, answer, confidence\}} and the integrator schema \texttt{\{analysis, final\_answer, reasoning\}}. Only the answer interface changes across tasks. LiveCodeBench uses code-generation fields, GPQA expects one option letter, and multimodal benchmarks pass the associated image with the text question. This design keeps the controller fixed while preserving each benchmark's native answer format and grading protocol.

\paragraph{Runtime behavior.}
At runtime, the controller maintains the candidate pool and repeatedly calls \explore{} until the exploration budget is exhausted or the controller decides to stop and synthesize. The same implementation supports both text-only and image-conditioned tasks, and all runs can optionally record the controller trace and sub-model outputs for later inspection. These choices make the evaluation pipeline inspectable without changing the core decision process.

\paragraph{Grading.}
LiveCodeBench uses deterministic grading via code execution. GPQA-Diamond uses exact option-letter matching, consistent with the official GPQA evaluation protocol~\citep{gpqa_paper}. HLE-Verified always uses an LLM judge (Claude Haiku 4.5) that compares the predicted answer against the reference answer. BabyVision uses hybrid grading: multiple-choice items are graded by string match on the extracted option letter, and fill-in-the-blank items use the same Haiku judge.

\paragraph{Cost accounting.}
All costs reported as \$/q reflect API token charges only, computed from per-model pricing and recorded token usage. For \our{}, this includes orchestrator tokens, explorer tokens, and (when applicable) integrator tokens. For reward-model reranking, the reported cost covers the $N{=}8$ candidate generation calls; the reward model itself runs locally and incurs no API cost but does require GPU compute not reflected in the dollar figure. Grading costs (LLM judge calls for HLE and BabyVision) are excluded from all rows. In all tables, \textbf{bold} marks the best accuracy per group and the lowest \$/q among test-time scaling methods.

\paragraph{Baseline implementations.}
All baselines use the same backbone model and the same benchmark-specific answer format as \our{}, so the comparison isolates the inference-time strategy.

\emph{Pass@1} generates a single independent response per question and returns its answer directly.

\emph{Majority Voting} draws \(N=8\) independent responses and selects the most frequent answer.

\emph{Self-Refine}~\citep{self_refine_paper} generates an initial draft, then enters a feedback-revision loop: a separate feedback call critiques the current draft and returns a correctness judgment; if the draft is judged incorrect, a revision call produces an updated draft conditioned on all prior drafts and feedbacks. The loop terminates when feedback returns a positive judgment or the iteration budget is exhausted. In practice, the feedback model frequently accepts the initial draft, so Self-Refine terminates after a single feedback call on the majority of questions.

\emph{Self-Refine (no early stop)} matches the default fixed-iteration setting of the original Self-Refine algorithm~\citep{self_refine_paper} (\S 3.3), which runs Feedback-Refine for a fixed number of rounds rather than relying on critic-decided early termination. This variant uses the identical control flow as our Self-Refine baseline, but changes the critic step in two ways. First, the critic prompt applies a skeptical review rubric: it asks the model to set the proposed reasoning aside, form a fresh intuition before re-engaging with the draft, and then answer five probing questions about the weakest step, hidden assumptions, and unused information in the question. Second, the critic is required to mark the current draft as not yet acceptable until the trajectory has produced at least two refined drafts beyond the initial generation, i.e.\ at least three total drafts. These constraints are prompt-level rather than hard-coded; Figure~\ref{fig:explore-distribution-all} reports the realized iteration-count distribution.

\emph{Budget Forcing}~\citep{s1_budget_forcing_paper} runs \(N\) sequential rounds on the same question. Round~1 generates an initial solution. Each subsequent round feeds the previous round's full reasoning trace back to the model with ``Wait'' appended, suppressing premature termination and forcing the model to continue deliberating before producing a new answer. All \(N\) rounds run unconditionally, and the final round's answer is returned. Because continuation rounds extend the existing reasoning rather than solving from scratch, they tend to produce shorter completions and lower per-round cost.

\emph{Reward-model reranking} (Skywork-Reward-V2 for text benchmarks, VisualPRM for multimodal benchmarks) draws the same \(N=8\) independent candidates, then scores each with a separately trained reward model and returns the highest-scoring answer. Skywork-Reward-V2 is an outcome reward model that assigns a single scalar score per response; VisualPRM is a process reward model that aggregates per-step scores. The reward models run locally and incur no API cost, but the candidate generation cost is identical to that of Majority Voting.

\emph{LLM Selection} draws the same \(N=8\) independent candidates as Majority Voting and reward-model reranking, then feeds all eight candidates back to the backbone model in a single call and asks it to analyze, compare, and synthesize a final answer. Candidates are generated upfront rather than chosen sequentially by a controller, so any improvement over Pass@1 is attributable to single-shot synthesis rather than to closed-loop stopping decisions. This baseline therefore isolates the contribution of LLM-based aggregation from adaptive sequential exploration; results are reported in Table~\ref{tab:explore-augmentation} alongside \attsmm{}, since both consume a fixed eight-explore budget per question.

\section{Qualitative Analysis}
\label{app:case-studies}

To understand \emph{how} \our{} improves over Pass@1, we examine the GPQA-Diamond questions where \our{} produces a correct answer but the first exploration does not. Two mechanisms appear. \emph{Majority correction}: the first explore gives the wrong answer, but subsequent explores converge on the correct option through independent analyses, and the orchestrator follows the consensus (Listing~\ref{lst:majority-case}). \emph{Minority extraction}: the explore majority is incorrect, but the orchestrator identifies a reasoning flaw in the majority and selects the correct minority candidate (Listing~\ref{lst:minority-case}).

\begin{promptbox}[label=lst:majority-case]{Majority correction on a Diels-Alder/NMR question (GPQA-Diamond)}Question: A dicarboxylic acid containing a cis-alkene was dehydrated to
the corresponding anhydride [...] reacted with 1,2,3,4-tetramethyl-
cyclopentadiene [...] How many peaks are in the CHCH3 region?
Options: (A) 2 peaks  (B) 3 peaks  (C) 4 peaks  (D) 8 peaks
Gold: A

Turn 1
  Thought:     trajectory empty; need a first candidate
  Action:      explore[]
  Observation: c_1 = (answer=D, confidence=0.85,
                     approach="Diels-Alder stereochemistry; 8 peaks")

Turn 2
  Thought:     single candidate; insufficient evidence
  Action:      explore[]
  Observation: c_2 = (answer=A, confidence=0.88,
                     approach="endo bicyclic product; symmetry collapses
                              methyls to bridge (6H) + ring (6H), 2 peaks")

Turn 3
  Thought:     candidates split 1-1; continue
  Action:      explore[]
  Observation: c_3 = (answer=A, confidence=0.88,
                     approach="independent confirmation via endo selectivity")

[Turns 4-5 produce two more A candidates; tally A=4, D=1.]

Turn 6
  Thought:     (*@\hl{four independent candidates converge on A through consistent}@*)
               (*@\hl{endo-selectivity reasoning;}@*) the dissenting D candidate assumed
               exo. Reasoning quality and vote count both favor A.
  Action:      stop[answer=A, confidence=0.75]
\end{promptbox}

\begin{promptbox}[label=lst:minority-case]{Minority extraction on a molecular-symmetry question (GPQA-Diamond)}Question: Which of the following molecules has C3h symmetry?
(A) Triisopropyl borate  (B) Triphenylene trianhydride
(C) Benzene trianhydride  (D) Quinuclidine
Gold: B

Turn 1
  Thought:     trajectory empty
  Action:      explore[]
  Observation: c_1 = (answer=A, confidence=0.92,
                     approach="BO3 plane gives sigma_h; C3h")

Turn 2
  Thought:     single candidate; need independent confirmation
  Action:      explore[]
  Observation: c_2 = (answer=B, confidence=0.85,
                     approach="isopropyl propeller breaks sigma_h;
                              A is C3, not C3h")

Turn 3
  Thought:     1-1 split with directly contradicting structural claims
  Action:      explore[]
  Observation: c_3 = (answer=A, confidence=0.88,
                     approach="repeats sp2-planar argument for A")

Turn 4
  Thought:     A leads 2-1 but the B candidate raised a specific topology
               issue not yet addressed; continue
  Action:      explore[]
  Observation: c_4 = (answer=B, confidence=0.82,
                     approach="C has perpendicular C2 axes -> D3h;
                              B's fused-ring inequivalence -> C3h")

[Turns 5-7 produce two more A candidates and one more B; tally A=4, B=3.]

Turn 8
  Thought:     majority favors A but every A candidate uniformly assumes
               sigma_h; (*@\hl{c\_2 and c\_4 explicitly identify the propeller-}@*)
               (*@\hl{conformation flaw.}@*) Reasoning quality overrides vote count.
  Action:      stop[answer=B, confidence=0.62]
\end{promptbox}

\section{Benchmark Details}
\label{app:benchmark-details}

This appendix details each benchmark's official evaluation protocol and our adaptation for Table~\ref{tab:main-results}. The top-group rows in that table cite the top-5 zero-shot leaderboard reference per benchmark with backbones labeled beneath each cell; backbones differ across cells, so these rows are not directly comparable to the middle and bottom groups, which share a Claude Sonnet 4.6 backbone so that differences there isolate the inference-time strategy.

\paragraph{HLE-Verified.}\label{app:hle-details}
We use the official HLE-Verified dataset release \texttt{skylenage/HLE-Verified} \citep{hle_verified_dataset_hf}, which partitions the 2,500 audited HLE items into Gold (668), Revision (1,143), and Uncertain (689) subsets \citep{hle_verified_paper}. HLE-Verified preserves the original HLE answer-based evaluation semantics: each item contains a problem statement, a final answer, and sometimes a rationale, but official scoring is driven by the problem and final answer, with rationale used only as auxiliary evidence during verification and revision \citep{hle_paper_nature,hle_verified_paper}. In the HLE-Verified evaluation paper, the reported baselines are run on the text-only portion of HLE using the official HLE system prompt, each model's default recommended decoding configuration, and five independent rollouts per item; the paper reports both \emph{avg5 Accuracy} and \emph{Calibration Error}, with calibration computed by the official HLE evaluation code \citep{hle_verified_paper}. In this paper, HLE reporting uses the first $n{=}100$ items of the 668-item Gold subset, while Table~\ref{tab:main-results} reproduces the official revised-subset accuracy references from \citet{hle_verified_paper} for context. Our grading uses an LLM-as-judge pipeline with Claude Haiku 4.5 as the judge model, which determines semantic equivalence between the predicted and gold answers.

\paragraph{LiveCodeBench.}\label{app:lcb-details}
We use the LiveCodeBench code generation release \texttt{livecodebench/code\_generation\_lite} \citep{lcb_dataset_hf}, configuration \texttt{v6} \citep{livecodebench_paper}. LiveCodeBench is a continuously updated programming benchmark sourced from LeetCode, AtCoder, and CodeForces. Each problem is tagged by release date, so official evaluations can be restricted to post-cutoff time windows \citep{livecodebench_paper}. In the official code generation setting, the model receives the natural-language problem statement together with example input-output pairs and must return a complete executable program. Correctness is determined purely by functional correctness on unseen tests, using the benchmark harness rather than an LLM judge \citep{livecodebench_paper}. The benchmark paper reports \emph{Pass@1} as the primary metric and also reports difficulty-stratified \emph{Pass@1} using the platform-provided Easy/Medium/Hard labels. The public generation leaderboard \citep{livecodebench_leaderboard} follows this same framing, and Table~\ref{tab:main-results} reports overall Pass@1 from the same leaderboard view. In this paper, LCB reporting uses the first $n{=}175$ problems of the v6 release. Our grading uses the official \texttt{lcb\_runner} execution harness with a 10-second per-test-case timeout.

\paragraph{BabyVision.}\label{app:babyvision-details}
We use the BabyVision dataset release \texttt{UnipatAI/BabyVision} \citep{babyvision_dataset_hf}, which contains 388 image-question pairs spanning four task families: Fine-grained Discrimination (163), Visual Tracking (83), Spatial Perception (91), and Visual Pattern Recognition (51) \citep{babyvision_paper}. The benchmark contains 135 multiple-choice items and 253 fill-in-the-blank items; in the official inference setup, each model is queried with a unified prompt template of the form ``\texttt{\{question\} Think about the question and give your final answer in \{format\}.}'', which standardizes answer formatting for downstream extraction and judging \citep{babyvision_paper}. Official scoring uses an LLM-as-judge pipeline: given a model answer and the gold answer, Qwen3-Max determines semantic equivalence under the benchmark's fixed judge prompt, which the authors report as having perfect agreement with their human evaluators on the benchmark responses they audited \citep{babyvision_paper}. The primary metric is \emph{Avg@3}, defined in the paper as the mean \emph{Pass@1} accuracy across three random runs, and the official tables additionally report the corresponding standard deviation; Table~\ref{tab:main-results} keeps only the overall mean accuracy in the main text. Our grading is hybrid: multiple-choice items (135 of 388) are graded by string match on the extracted option letter, while fill-in-the-blank items (253 of 388) use an LLM-as-judge pipeline with Claude Haiku 4.5, following the official semantic-equivalence protocol with the substituted judge model.

\paragraph{GPQA-Diamond.}\label{app:gpqa-details}
We use the GPQA release \texttt{Idavidrein/gpqa} \citep{gpqa_dataset_hf}, configuration \texttt{gpqa\_diamond}, which corresponds to the 198-question Diamond subset introduced in the GPQA paper \citep{gpqa_paper}. GPQA is a graduate-level four-option multiple-choice science benchmark spanning biology, physics, and chemistry, and GPQA-Diamond is the highest-stringency subset retained after expert-accuracy and non-expert-difficulty filtering \citep{gpqa_paper}. The official task is closed-book answer selection unless otherwise stated: the model is shown the question and four candidate options and must output exactly one final option. Scoring is exact option-key accuracy with no semantic judge and a random-guess baseline of \(25\%\) \citep{gpqa_paper}. In addition to Diamond-set accuracy, the original GPQA paper also reports accuracies on the Extended and Main sets and distinguishes closed-book few-shot/CoT baselines from an open-book GPT-4-with-search baseline; Table~\ref{tab:main-results} reports the Diamond-set accuracy column used by current public references. Our grading extracts the option letter (A--D) from the model response via regex and performs exact matching against the gold option key, with no LLM judge, consistent with the official protocol.

\paragraph{RBench-V.}\label{app:rbenchv-details}
We use the RBench-V dataset release at \citet{rbenchv_dataset_hf}. The benchmark consists of 803 multi-modal reasoning problems spanning four task families: Math (176), Physics (157), Counting (195), and Games (275) \citep{rbenchv_paper}. Every item is image-conditioned, and the question pool mixes multiple-choice items with open-ended free-form answers. Because the benchmark includes both multiple-choice and open-ended questions, the official protocol grades all items through a unified LLM-as-judge framework with GPT-4o as the judge model, reporting Top-1 accuracy as the primary metric \citep{rbenchv_paper}. Our grading reuses an LLM-as-judge pipeline with Claude Haiku 4.5 in the judge role, preserving the open-ended semantic-equivalence semantics required by the official protocol. The official RBench-V protocol additionally scores intermediate image generations (e.g., auxiliary lines drawn onto the input figure); the orchestrators and explorers used in this paper do not natively emit raster images, so our explorers receive only the original problem (image plus text) without ground-truth answers and reason about the visual content through textual descriptions delegated by the orchestrator.
\section{Results on RBench-V}
\label{app:rbenchv-results}

\begin{table}[!htb]
\centering
\caption{Math accuracy on the RBench-V Math subset ($n{=}176$).}
\label{tab:rbenchv-math-prelim}
\scriptsize
\setlength{\tabcolsep}{3pt}
\begin{tabular}{l r}
\toprule
 & Math (\%) \\
\midrule
\groupheader{2}{Zero-shot models}
OpenAI o3 & 48.3 \\
OpenAI o4-mini & 43.2 \\
Gemini 2.5 Pro & 42.6 \\
Doubao-1.5-thinking-pro-m & 38.6 \\
OpenAI o1 & 34.7 \\
\midrule
\groupheader{2}{Test-time scaling methods}
Pass@1 & 54.5 \\
\rowcolor{ClaudeOrange!8} \our{} & \textbf{61.4} \\
\bottomrule
\end{tabular}
\end{table}

Table~\ref{tab:rbenchv-math-prelim} reports \our{} and Pass@1 on the RBench-V Math subset ($n{=}176$) under a Gemini 3 Flash Preview backbone. The zero-shot rows are cited from Table~3 of \citet{rbenchv_paper}; the bottom group reports Pass@1 and \our{} under our own evaluation, with the Gemini 3 Flash Preview backbone serving as orchestrator and explorer end-to-end. \our{} reaches $61.4\%$, ahead of Pass@1 by $6.9$ points; both rows exceed the highest cited zero-shot reference (OpenAI o3, $48.3\%$).

\section{Full Per-Split Results}
\label{app:full-results}

This appendix reports the per-split breakdowns summarized in Table~\ref{tab:main-results} (\S\ref{sec:main-results}). Table~\ref{tab:full-per-split-main} reproduces the LiveCodeBench Easy/Medium/Hard split and the BabyVision Fine-grained / Tracking / Spatial / Pattern split for the four-benchmark subset.

\begin{table}[H]
\centering
\caption{Per-split breakdown for the main four benchmarks (HLE-Verified, LiveCodeBench, GPQA-Diamond, BabyVision). Each subtable shows the zero-shot leaderboard references (upper group) and test-time scaling methods (lower group). \textbf{Bold} marks the best per group.}
\label{tab:full-per-split-main}
\scriptsize
\setlength{\tabcolsep}{2pt}
\begin{minipage}[t]{0.38\textwidth}
\centering
\textit{(a) HLE-Verified}\\[3pt]
\begin{tabular}{l r}
\toprule
 & Acc. (\%) \\
\midrule
\groupheader{2}{Zero-Shot Models}
Gemini 3 Pro & 48.93 \\
GPT-5.2-High & \textbf{52.48} \\
Claude Opus 4.5 & 48.16 \\
Grok 4.1 Fast Reasoning & 43.07 \\
Claude Opus 4.6 & 50.16 \\
DeepSeek-V3.2 & 47.45 \\
Qwen3-Max-Thinking & 48.48 \\
\midrule
\groupheader{2}{Test-Time Scaling Methods}
Pass@1 & 48.00 \\
Self-Refine & 53.00 \\
Self-Refine (no early stop) & 58.00 \\
Budget Forcing & 51.00 \\
Skywork-Reward-V2 & 52.00 \\
\our{} & 56.00 \\
\attsmm{} & \textbf{60.00} \\
\bottomrule
\end{tabular}
\end{minipage}\hfill%
\begin{minipage}[t]{0.60\textwidth}
\centering
\textit{(b) LiveCodeBench}\\[3pt]
\begin{tabular}{l r r r r}
\toprule
 & Pass@1 & Easy & Med. & Hard \\
\midrule
\groupheader{5}{Zero-Shot Models}
O4-Mini (High) & \textbf{80.2} & \textbf{99.1} & \textbf{89.4} & \textbf{63.5} \\
O3 (High) & 75.8 & \textbf{99.1} & 84.4 & 57.1 \\
O4-Mini (Med.) & 74.2 & 98.2 & 86.5 & 52.7 \\
Gemini-2.5-Pro & 73.6 & \textbf{99.1} & 87.2 & 50.2 \\
DeepSeek-R1 & 73.1 & 98.7 & 85.2 & 50.7 \\
\midrule
\groupheader{5}{Test-Time Scaling Methods}
Pass@1 & 77.14 & 95.35 & 88.46 & 60.00 \\
Self-Refine & 80.57 & \textbf{100.00} & 90.38 & 63.75 \\
Self-Refine (no early stop) & 82.29 & \textbf{100.00} & 90.38 & 67.50 \\
Budget Forcing & 80.00 & \textbf{100.00} & 88.46 & 63.75 \\
Skywork-Reward-V2 & 78.86 & \textbf{100.00} & 90.38 & 60.00 \\
\our{} & 82.29 & \textbf{100.00} & 90.38 & 67.50 \\
\attsmm{} & \textbf{85.63} & \textbf{100.00} & \textbf{92.31} & \textbf{73.42} \\
\bottomrule
\end{tabular}
\end{minipage}

\vspace{6pt}

\begin{minipage}[t]{0.38\textwidth}
\centering
\textit{(c) GPQA-Diamond}\\[3pt]
\begin{tabular}{l r}
\toprule
 & Acc. (\%) \\
\midrule
\groupheader{2}{Zero-Shot Models}
Gemini 3.1 Pro Preview & \textbf{94.1} \\
GPT-5.4 (xhigh) & 92.0 \\
GPT-5.3 Codex (xhigh) & 91.5 \\
Gemini 3 Pro Preview (high) & 90.8 \\
GPT-5.2 (xhigh) & 90.3 \\
\midrule
\groupheader{2}{Test-Time Scaling Methods}
Pass@1 & 83.33 \\
Majority Voting & 83.33 \\
Self-Refine & 82.83 \\
Self-Refine (no early stop) & 85.35 \\
Budget Forcing & 83.84 \\
Skywork-Reward-V2 & 83.84 \\
\our{} & 85.86 \\
\attsmm{} & \textbf{88.38} \\
\bottomrule
\end{tabular}
\end{minipage}\hfill%
\begin{minipage}[t]{0.60\textwidth}
\centering
\textit{(d) BabyVision}\\[3pt]
\begin{tabular}{l r r r r r}
\toprule
 & Overall & Fine-gr. & Tracking & Spatial & Pattern \\
\midrule
Human & 94.1 & 92.3 & 94.6 & 94.7 & 97.8 \\
\midrule
\groupheader{6}{Zero-Shot Models}
Gemini3-Pro-Preview & \textbf{49.7} & \textbf{46.2} & \textbf{43.4} & \textbf{53.7} & 53.9 \\
GPT-5.2 & 34.4 & 27.3 & 34.9 & 35.2 & \textbf{54.9} \\
Doubao-1.8 & 30.2 & 39.2 & 15.7 & 24.7 & 37.7 \\
Qwen3-VL-Plus & 19.2 & 21.8 & 11.5 & 18.1 & 25.5 \\
Grok-4 & 16.2 & 11.0 & 24.1 & 13.2 & 24.8 \\
\midrule
\groupheader{6}{Test-Time Scaling Methods}
Pass@1 & 19.59 & 20.86 & 19.28 & 19.78 & 15.69 \\
Majority Voting & 23.20 & \textbf{23.93} & 21.69 & 23.08 & 23.53 \\
Self-Refine & 21.39 & 20.25 & 22.89 & 20.88 & 23.53 \\
Self-Refine (no early stop) & 21.13 & 18.40 & 25.30 & 20.88 & 23.53 \\
Budget Forcing & 21.91 & 20.86 & \textbf{28.92} & 21.98 & 13.73 \\
VisualPRM & 20.86 & 20.86 & 21.69 & 21.98 & 17.65 \\
\our{} & 23.71 & 22.09 & 24.10 & \textbf{26.37} & 23.53 \\
\attsmm{} & \textbf{23.97} & 22.09 & 27.71 & 23.08 & \textbf{25.49} \\
\bottomrule
\end{tabular}
\end{minipage}
\end{table}

\section{Explore-Count Distributions}
\label{app:explore-distributions}

\begin{figure}[H]
\centering
\includegraphics[width=\textwidth]{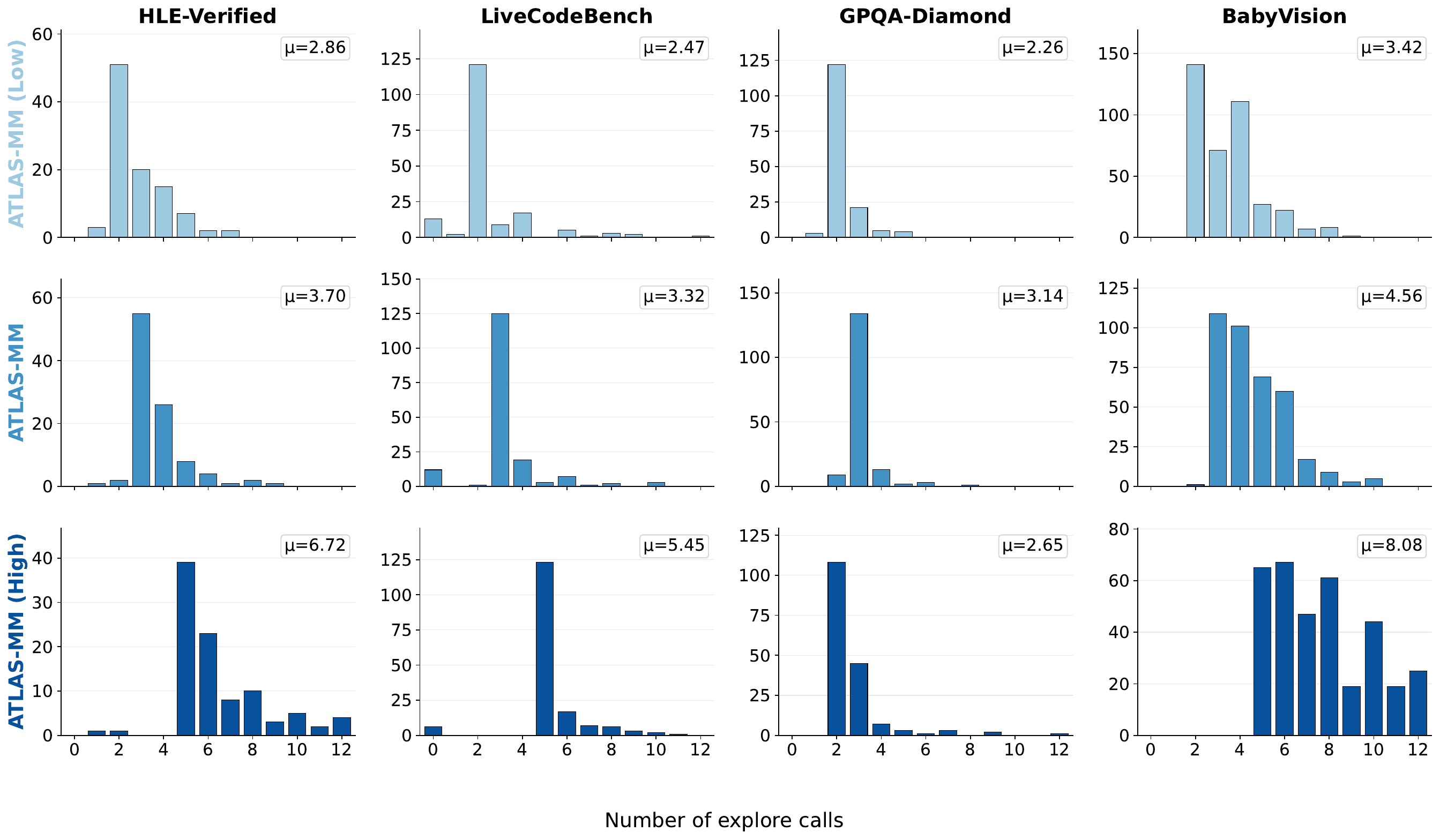}
\caption{Per-question explore-call distributions for \attsmm{} at three explore-effort levels (rows: Low / Medium / High) across four benchmarks (columns). Effort tier rises top-to-bottom. Same data source as the main-text KDE summary in Figure~\ref{fig:kde-effort-ablation}.}
\label{fig:bars-effort-ablation}
\end{figure}

This appendix reports the per-question explore-call distributions underlying the two main-text KDE summaries in Figures~\ref{fig:kde-effort-ablation} and~\ref{fig:kde-method-comparison}. Figures~\ref{fig:bars-effort-ablation} and~\ref{fig:explore-distribution-all} below are matrices of discrete histograms: rows correspond to methods, columns to benchmarks. Each panel's $x$-axis is the number of explore calls used for one question, and the $y$-axis is the number of questions at that count; bar heights are annotated with exact counts, and the boxed annotation reports the sample size $n$ and per-question mean $\mu$.

\subsection{\attsmm{} under varying explore effort}
\label{app:explore-distributions-effort}

\noindent\textit{(i)~Low effort.} All four benchmarks concentrate mass at $n{\in}\{2,3\}$ (HLE: $51$ at $n{=}2$; LCB: $121$ at $n{=}2$; GPQA: $142$ at $n{=}2$; BV: $141$ at $n{=}2$ plus $111$ at $n{=}3$), establishing a baseline mode of two-to-three explores per question.

\noindent\textit{(ii)~Medium effort.} The mode shifts up by one on HLE/LCB/GPQA (from $n{=}2$ to $n{=}3$, with $55${--}$154$ trajectories at the new mode), while BV develops a wider plateau between $n{=}3$ and $n{=}4$ ($109$ and $101$ trajectories).

\noindent\textit{(iii)~High effort.} HLE, LCB, and BV all exhibit a marked rightward shift: HLE's mode moves to $n{=}5$ ($39$ trajectories) with appreciable mass through $n{=}10$; LCB's mode moves to $n{=}5$ ($123$ trajectories) with a small tail; BV flattens dramatically between $n{=}5$ and $n{=}12$ ($\geq 19$ trajectories in every bin).

\noindent\textit{(iv)~GPQA-Diamond.} The High-effort distribution returns mass to $n{=}2$ ($112$ trajectories, $\approx 56.6\%$, up from $12$ trajectories at Medium), with secondary mass at $n{=}3$ ($59$ trajectories). The mean drops from $\mu{=}3.19$ at Medium to $\mu{=}2.75$ at High. This distributional contraction is the direct cause of the Medium$\to$High cost trajectory ($+10.5\%$, versus $1.5${--}$1.8\times$ on the free-form benchmarks) reported in Figure~\ref{fig:effort-ablation} and discussed in \S\ref{sec:effort-ablation}.

\subsection{\our{} vs.\ Self-Refine baselines}
\label{app:explore-distributions-method}

\begin{figure}[H]
\centering
\includegraphics[width=\textwidth]{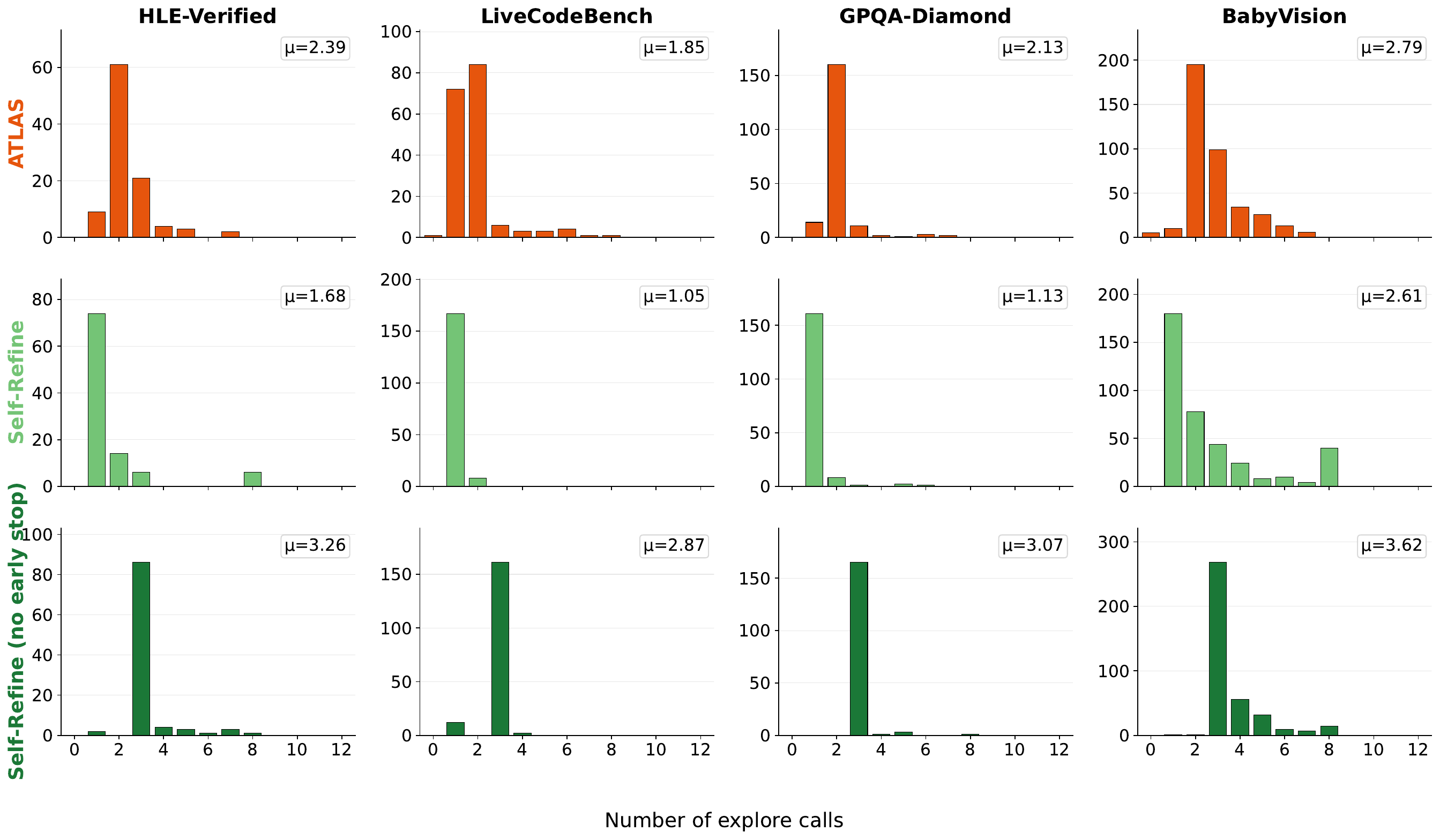}
\caption{Per-question explore-call distributions for \our{} (orange), Self-Refine (light green), and Self-Refine (no early stop) (dark green), all on the shared Sonnet 4.6 backbone (rows) across four benchmarks (columns). All three rows share the same exploration budget $T{=}8$, so differences in the observed distributions reflect each method's stopping mechanism. Same data source as the main-text KDE summary in Figure~\ref{fig:kde-method-comparison}. Budget Forcing terminates at the cap by construction, giving a degenerate distribution at $T{=}8$.}
\label{fig:explore-distribution-all}
\end{figure}

All four benchmarks share the same exploration budget $T{=}8$ and the same orchestrator prompt for the \our{} row, so differences in the observed distributions reflect the orchestrator's input-conditioned stopping decisions rather than externally imposed budgets. We highlight five observations from the \our{} row of this figure, then return to the contrast with Self-Refine.

\paragraph{Observation 1: Mean explore count tracks benchmark difficulty.}
The per-benchmark mean \explore{} count for \our{} ranges from $1.85$ on LiveCodeBench to $2.79$ on BabyVision, with GPQA-Diamond at $2.14$ and HLE-Verified at $2.39$. Harder benchmarks (those on which test-time scaling methods reach lower absolute accuracy in Table~\ref{tab:main-results}) receive more exploration per question on average. Given that all four benchmarks share the same budget and the same orchestrator prompt, this is a direct numerical expression of the adaptive compute allocation claim: the controller spends more calls where first-pass evidence is less conclusive.

\paragraph{Observation 2: The modal behavior is ``explore twice,'' not ``explore once.''}
Three of the four benchmarks (HLE-Verified, GPQA-Diamond, BabyVision) place their \our{} distribution mode at $n{=}2$ \explore{} calls, and the share of questions stopping at $n{=}1$ is below $15\%$ on each of these three benchmarks. LiveCodeBench is the only exception, with a bimodal distribution in which the first mode at $n{=}1$ ($72$ trajectories) is comparable in size to the second mode at $n{=}2$ ($84$). The default orchestrator behavior is therefore ``explore twice before synthesizing,'' not ``explore once and submit.'' The contrast with Self-Refine is sharp: under Self-Refine (Figure~\ref{fig:explore-distribution-all}, middle row) the mode is $n{=}1$ on three of four benchmarks (mean $1.05$--$1.68$ on HLE/LCB/GPQA) because the critic accepts the initial draft on the first revision check for the vast majority of questions. \our{}'s mass is centered between Self-Refine's one-shot acceptance and a hypothetical fixed-budget schedule, reflecting that the orchestrator's stop decisions are conditioned on accumulated evidence rather than on a single one-shot critic check.

\paragraph{Observation 3: The budget cap $T{=}8$ is almost never reached.}
Across the $861$ questions in the four benchmarks combined, exactly one question consumes the full exploration budget of $T{=}8$ (on LiveCodeBench). Questions using $n \ge 6$ \explore{} calls remain rare on every benchmark: $6$ on LiveCodeBench, $6$ on GPQA-Diamond, $2$ on HLE-Verified, and $19$ on BabyVision. The orchestrator terminates exploration well before the cap on virtually every question, supporting the interpretation that $T$ functions as a safety ceiling rather than a target.

\paragraph{Observation 4: LiveCodeBench bimodality reflects per-problem difficulty recognition.}
LiveCodeBench is the only benchmark with a substantial mass at $n{=}1$ ($41\%$ of questions). These single-explore questions likely correspond to problems for which the first candidate is obviously sufficient: LiveCodeBench's Easy tier already reaches $100\%$ Pass@1 (Appendix~\ref{app:full-results}), and a fraction of Medium problems are similarly unambiguous once a first candidate is produced. The orchestrator's ability to terminate immediately on such problems, rather than always continuing to the modal $n{=}2$ seen on the other three benchmarks, is direct evidence that the controller performs per-problem difficulty recognition rather than following a fixed exploration schedule.

\paragraph{Observation 5: BabyVision is the inverse-pattern case.}
BabyVision is the only benchmark on which the highest mean \explore{} count ($2.79$) co-occurs with the lowest accuracy among the four ($23.71\%$ in Table~\ref{tab:main-results}). The orchestrator explores most aggressively on BabyVision (roughly half of all questions receive three or more \explore{} calls). This inverse pattern indicates that BabyVision approaches the capability ceiling of the explorer pool: more exploration uncovers more candidate disagreement but does not yield a correct answer that the orchestrator can extract.

\section{Prompt Families}\label{app:prompt-families}

The orchestrator's system prompt is composed at runtime from three orthogonal slots: a variant template that fixes the \explore{} action space, a stopping-threshold instruction, and a finalize instruction. Two of these slots reflect framework design choices that are realised purely at the prompt level rather than as runtime knobs. First, the three \explore{} action variants (\S\ref{sec:variants}) each carry a distinct template instead of sharing one template with a switch. Second, the explore-effort hyperparameter is itself a short stopping-threshold instruction injected into the prompt. We document the contents of each slot below; the few-shot example blocks that appear inside the templates are omitted, since their role is operational.

\subsection{Stopping-threshold instructions}

Explore-effort selects one of three short instructions. Each is injected verbatim into the system prompt and changes the orchestrator's default convergence criterion without touching any other component.

\begin{promptbox}{Low effort}Your stopping threshold leans loose. When two independent candidates agree on the same answer, that's typically enough for you to feel ready to submit -- chasing extra confirmation rarely repays the cost of another call. Clean convergence ends the round.
\end{promptbox}

\begin{promptbox}{Medium effort}Your stopping threshold is moderate. Two converging candidates earns the answer your trust, but on anything non-trivial you usually prefer one more cross-method explore before committing. A lone confident candidate seldom satisfies you, since self-reported confidence is unreliable.
\end{promptbox}

\begin{promptbox}{High effort}Your stopping threshold leans strict. On a hard problem, two agreeing candidates feel encouraging but not yet conclusive -- you typically prefer three independent methods agreeing, or an explicit cross-check (different model). When you're torn between submitting and exploring once more, you usually choose the explore.
\end{promptbox}

\subsection{Variant templates for the \explore{} action space}

The three \explore{} action variants (base \our{}, \attsmm{}, and \attsmi{}) each ship with a dedicated system-prompt template. We paste the three templates verbatim below, omitting only the worked-example sections at the bottom of each template (these contain few-shot dialogues used to teach the LLM the expected reasoning pattern). The placeholders \texttt{\{stop\_instruction\}} and \texttt{\{finalize\_instruction\}} are filled at runtime from the listings in the previous and next subsection respectively.

\begin{promptbox}{Base \protect\our{}}You are a meta-reasoning orchestrator. You manage a pool of candidate solutions for a problem.

After each explore, you will see the candidate result (answer, confidence, approach, reasoning).

{finalize_instruction}

{stop_instruction}

## Principles (HIGHEST PRIORITY -- override any urge to solve the problem yourself)

1. You cannot solve problems yourself. Your only window into the problem is what solvers return. Reasoning about the problem content -- analyzing algorithms, deriving formulas, writing code -- constitutes solving, even when framed as "analysis" or "synthesis". Any answer without candidate evidence is baseless and undermines the system.
2. A single candidate, regardless of its self-reported confidence, does not constitute sufficient evidence. Self-reported confidence is poorly calibrated.
3. Genuine convergence means independent solvers arriving at the same answer through different methods. Candidates that agree but share the same reasoning path may reflect a shared misconception rather than true convergence.
4. Solver failure (timeout, empty answer) reflects the problem's difficulty. Repeated failures reinforce this -- more attempts will not help. If solvers timed out, you will almost certainly fail too. When solvers consistently fail, the problem is practically unsolvable -- submitting an empty answer is better than attempting it yourself, because you would also fail and waste budget in the process.
5. CRITICAL: Each explore costs budget. Giving up is a valid and preferred action -- when candidates provide no useful information, you MUST submit an empty answer immediately rather than wasting budget or attempting to solve.

Every decision you make must be grounded in one of the principles above. Explicitly cite which principle justifies your action.
\end{promptbox}

\begin{promptbox}{\protect\attsmm{}}You are a meta-reasoning orchestrator. You manage a pool of candidate solutions for a problem.

You have access to multiple solver models of different capability and cost. Each explore call lets you choose which model to use.

After each explore, you will see the candidate result (answer, confidence, approach, reasoning), which model produced it, and how much it cost.

{finalize_instruction}

{stop_instruction}

## Model Profiles

Pricing (per million tokens): Haiku $1 input / $5 output, Sonnet $3 input / $15 output, Opus $5 input / $25 output.
Haiku is 3-5x cheaper than Sonnet per call. Sonnet is ~2x cheaper than Opus per call.

Public benchmark scores (from official Anthropic announcements):

| Benchmark | Category | Haiku | Sonnet | Opus |
|---|---|---|---|---|
| MATH-500 | Math (standard) | -- | 97.8% | 97.6% |
| AIME 2025 | Math (competition) | 80.7% | 83.0% | 99.8% |
| AIME 2026 | Math (competition) | -- | -- | 96.7% |
| GPQA-Diamond | Graduate science | 73.0% | 89.9% | 91.3% |
| HLE | Hard reasoning | -- | 33.2% | 40.0% |
| ARC-AGI-2 | Novel problem-solving | -- | 58.3% | 68.8% |
| MMMLU | General knowledge | 83.0% | 89.3% | 91.1% |
| SWE-bench Verified | Coding | 73.3% | 79.6% | 80.8% |
| Terminal-Bench 2.0 | Coding (agentic) | 41.0% | 59.1% | 65.4% |
| OSWorld-Verified | Computer use | 50.7% | 72.5% | 72.7% |
| BrowseComp | Web search | -- | 74.7% | 84.0% |

Use this table together with the pricing to decide which model to dispatch for each explore call.
\end{promptbox}

\begin{promptbox}{\protect\attsmm{} (continued)}## Principles (HIGHEST PRIORITY -- override any urge to solve the problem yourself)

1. You cannot solve problems yourself. Your only window into the problem is what solvers return. Reasoning about the problem content -- analyzing algorithms, deriving formulas, writing code -- constitutes solving, even when framed as "analysis" or "synthesis". Any answer without candidate evidence is baseless and undermines the system.
2. A single candidate, regardless of its self-reported confidence, does not constitute sufficient evidence. Self-reported confidence is poorly calibrated.
3. Genuine convergence means independent solvers arriving at the same answer through different methods. Candidates from different models that agree provide stronger evidence than candidates from the same model.
4. A weaker model failing does not mean a stronger model will also fail. Start with cheaper models first; escalate when they fail or disagree. Only when the strongest model fails repeatedly is the problem beyond reach.
5. CRITICAL: Each explore costs real money. Giving up is a valid and preferred action -- when candidates provide no useful information, you MUST submit an empty answer immediately rather than wasting budget or attempting to solve.

Before each explore call, explicitly reason about: (a) which model to use and why, citing cost data; (b) what you expect to learn from this call.

Every decision you make must be grounded in one of the principles above. Explicitly cite which principle justifies your action.
\end{promptbox}

\begin{promptbox}{\protect\attsmi{}}You are a meta-reasoning orchestrator. You manage a pool of candidate solutions for a problem.

Each explore call dispatches a fresh, independent solver. You may optionally pass `additional_prompt` (a short string) to steer the solver toward a specific direction -- a different solving method, a particular sub-step to focus on, or a stricter sanity check.

After each explore, you will see the candidate result (answer, confidence, approach, reasoning).

{finalize_instruction}

{stop_instruction}

## When to use additional_prompt

Default: omit it. The first explore call should never use additional_prompt -- you have no candidates yet, so no specific direction is justified.

Use additional_prompt only when prior candidates give you concrete information that justifies a directional ask:
- Prior candidates all converged on the same answer via the same method. Ask the next solver to try a fundamentally different method to verify -- approach diversity is stronger evidence than method-redundant agreement (principle 3).
- Prior candidates disagree on a specific sub-step. Ask the next solver to focus on that sub-step.
- A candidate has high confidence but its reasoning has an obvious weak link. Ask the next solver to scrutinize that link.
- Prior candidates are all shallow or hand-wavy. Ask the next solver to be more rigorous on a named part.

When you do use additional_prompt, write it short and concrete. "Try a different method" is fine; "focus on whether step 5's case analysis is exhaustive" is better.

## Principles (HIGHEST PRIORITY -- override any urge to solve the problem yourself)

1. You cannot solve problems yourself. Your only window into the problem is what solvers return. Reasoning about the problem content -- analyzing algorithms, deriving formulas, writing code -- constitutes solving, even when framed as "analysis" or "synthesis". Any answer without candidate evidence is baseless.
2. A single candidate, regardless of self-reported confidence, does not constitute sufficient evidence. Self-reported confidence is poorly calibrated.
3. Genuine convergence means independent solvers arriving at the same answer through DIFFERENT methods. Same-method agreement may reflect a shared misconception. additional_prompt is your tool for forcing method diversity when the explore pool has drifted into method-redundancy.
4. Solver failure (timeout, empty answer) reflects the problem's difficulty. Repeated failures reinforce this -- more attempts will not help.
5. CRITICAL: Each explore costs budget. Giving up is valid -- when candidates provide no useful information, submit an empty answer immediately rather than wasting budget or attempting to solve.

Every decision must cite which principle justifies the action.
\end{promptbox}

\subsection{Finalize instructions}

The third slot decides whether the final answer is produced by the orchestrator itself or delegated to a separate synthesizer.

\begin{promptbox}{Without integrator (default)}Your final answer must be derived from candidate outputs. You may combine insights from multiple candidates, but you cannot introduce information that no candidate provided. If no candidate produced useful information, you should give up and submit an empty answer to save cost.
\end{promptbox}

\begin{promptbox}{With integrator}When you are ready, call `integrate` to dispatch a synthesizer that produces the final answer from all candidates.
\end{promptbox}

\end{document}